\documentclass{WileyMSP-template}
\usepackage[dvipsnames,table,xcdraw]{xcolor}
\usepackage{amsmath}
\usepackage{amsfonts}
\usepackage{amssymb}
\usepackage{graphicx}
\usepackage{subcaption}
\usepackage{cite}

\usepackage{enumitem}
\usepackage{booktabs}
\usepackage{multirow}
\usepackage[hyphens]{url}
\usepackage[hidelinks]{hyperref}
\usepackage{soul,color}
\DeclareSymbolFont{matha}{OML}{txmi}{m}{it}
\DeclareMathSymbol{\varv}{\mathord}{matha}{118}
\DeclareMathOperator{\hyab}{HyAB}
\usepackage{bbm}

\DeclareRobustCommand{\revision}[1]{{\sethlcolor{white}\hl{#1}}}

\usepackage{soul}

\begin{document}

\pagestyle{fancy}
\rhead{\includegraphics[width=2.5cm]{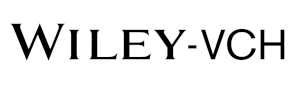}}

\title{
\revision{THOR2: Topological Analysis for 3D Shape and Color-Based Human-Inspired Object Recognition in Unseen Environments}}
\maketitle


\author{Ekta U. Samani}
\author{Ashis G. Banerjee*}


\dedication{}


\begin{affiliations}


Dr. Ekta U. Samani\\
Department of Mechanical Engineering\\
University of Washington, Seattle, WA 98195, USA\\

Prof. Ashis G. Banerjee \\
Department of Industrial \& Systems Engineering and the Department of Mechanical Engineering\\
University of Washington, Seattle, WA 98195, USA\\
ashisb@uw.edu

\end{affiliations}



\begin{abstract}
Visual object recognition in unseen and cluttered indoor environments is a challenging problem for mobile robots. This study presents a 3D shape and color-based descriptor, TOPS2, for point clouds generated from RGB-D images and an accompanying recognition framework, THOR2. The TOPS2 descriptor embodies object unity, a human cognition mechanism, by retaining the slicing-based topological representation of 3D shape from the TOPS descriptor \cite{samani2024persistent} while capturing object color information through slicing-based color embeddings computed using a network of coarse color regions. These color regions, analogous to the MacAdam ellipses identified in human color perception, are obtained using the Mapper algorithm, a topological soft-clustering technique. THOR2, trained using synthetic data, demonstrates markedly improved recognition accuracy compared to THOR, its 3D shape-based predecessor, on two benchmark real-world datasets: the OCID dataset capturing cluttered scenes from different viewpoints and the UW-IS Occluded dataset reflecting different environmental conditions and degrees of object occlusion recorded using commodity hardware. THOR2 also outperforms baseline deep learning networks, and a widely-used \revision{Vision Transformer} (ViT) adapted for RGB-D inputs \revision{trained using synthetic and limited real-world data} on both the datasets. Therefore, THOR2 is a promising step toward achieving robust recognition in low-cost robots.

\end{abstract}





\section{Introduction}

Object recognition plays a crucial role in robot visual perception, as it supports a wide range of tasks that industrial and domestic use robots must perform, such as manipulation of objects. Object recognition in humans is remarkably advanced; people can identify a wide array of objects in diverse settings, overcoming occlusion-related challenges or changes in appearance, viewpoint, size, scale, or pose. \revision{People can also distinguish between individual instances of objects, as, for example, a softball from a tennis ball, even though both are balls.} As a step toward achieving comparable performance in robots 
with commodity hardware, our previous work \cite{samani2024persistent} presents a 3D shape-based recognition framework, THOR, inspired by a human cognition mechanism known as \textit{object unity} \cite{goldstein2016sensation}. In this article, we extend our work to propose the THOR2 framework, which incorporates object color information. Analogous to the \textit{MacAdam ellipses} \cite{macadam1942visual} that represent colors indistinguishable to an average human, in a first-of-its-kind attempt, we obtain coarse color regions that are considered equivalent to the robot. We obtain these regions using a topological soft-clustering technique known as the Mapper algorithm \cite{singh2007topological} and use them to compute slicing-based object color embeddings. These embeddings are interleaved with our previously proposed 3D shape-based descriptor to obtain a 3D shape and color-based descriptor.


The primary challenges in developing an object \revision{instance} recognition method for low-cost robots used in everyday scenarios involve maintaining consistent performance regardless of varying environmental conditions (such as illumination and background), sensor quality, and the level of clutter in the surroundings. Early deep learning models \cite{ren2015faster, liu2016ssd, redmon2016you} achieve impressive performance in domain-specific tasks but experience performance degradation in unseen domains with different environmental conditions \cite{samani2021visual}. Domain adaptation and 
 generalization-based variants of such models have been explored. However, domain adaptation methods require label-agnostic data from the target domain for model training, and domain generalization methods require abundant real-world training data, which poses barriers to deploying such methods on robots with commodity hardware \cite{antonik2019human}. \revision{Therefore, we consider an alternative paradigm: we consider a single synthetic source domain to obtain the descriptors suitable for object instance recognition across multiple real-world target domains.}

Our previous work \cite{samani2024persistent} proposes a topological descriptor computed from depth images, known as TOPS (i.e., \underline{T}opological features \underline{O}f \underline{P}oint cloud \underline{S}lices). Persistent homology is applied to slicing-style filtrations of simplicial complexes constructed from the object's point cloud to obtain the descriptor. Our approach ensures similarities between the descriptors of the occluded and the corresponding unoccluded objects, embodying object unity. When used with the human-inspired recognition framework, THOR (i.e., \underline{T}OPS for \underline{H}uman-inspired \underline{O}bject \underline{R}ecognition), the TOPS descriptor shows promising robustness to occlusions. However, recognition using 3D shape information alone is challenging. Additional information cues, such as the color of an object, help achieve improved recognition \cite{lai2011large,kasaei2021investigating}, and are particularly crucial to distinguish among the \revision{objects instances} with similar geometry.

Several hand-crafted representations \cite{lai2011large,bo2013unsupervised}, multimodal convolutional neural networks \cite{gao2019rgb}, recurrent networks \cite{loghmani2019recurrent,caglayan2022cnns}, and transformer-based approaches \cite{tziafas2023early, xiong2023enhancing} for recognition using shape and color have been proposed. However, our paradigm requires color representations that transfer well from simulation to the real world. Obtaining such representations is challenging because the observed chromaticity of objects varies with the lighting conditions \cite{corke2023robotics}. The image signal processors on current RGB-D cameras run automatic white-balancing algorithms to remove undesirable color casts caused by environmental lighting. However, such algorithms require further work to completely tune out the effect of illumination conditions in typical real-world scenes \cite{afifi2022auto}. Therefore, we use the coarse color regions identified using topological soft clustering instead of specific colors to compute the color representations for supporting object recognition. The key contributions of our work are:

\begin{itemize}[leftmargin=*]

    \item We identify color regions (clusters of similar colors) in the standard RGB color space using the Mapper algorithm \cite{singh2007topological} and capture their connectivity in a color network.
    \item We propose a color network-based computation of color embeddings to obtain the TOPS2 descriptor for 3D shape and color-based recognition of occluded objects using an accompanying human-inspired framework THOR2.
    \item We show that THOR2 outperforms deep learning-based end-to-end models trained with synthetic data, including a widely-used transformer network adapted for RGB-D object recognition in unseen cluttered environments.
\end{itemize}


\section{Related Work}

Several approaches have been proposed for shape and color-based object recognition using RGB-D images, ranging from handcrafted descriptor-based methods to learning-based methods. In the following sections, we look at existing approaches for obtaining shape and color-based object representations from RGB-D images for recognition.

\subsection{Hand-crafted Representations:}
Most early works in RGB-D object recognition adopt the following framework. First, features capturing the visual appearance of objects are computed from the RGB image. Next, features capturing the object shape are computed from the corresponding depth image.  Often, more than one type of color-based features and shape-based features are computed and fused into a global descriptor through concatenation \cite{paulk2014supervised} or aggregate representations \cite{browatzki2011going, lai2011large}. These global descriptors are then used for recognition using a classifier. For instance, Lai et al. \cite{lai2011large} compute multiple features such as SIFT, color histograms, and texton histograms to capture the visual appearance of objects. Spin images are then computed from the depth image to capture the shape of objects. Efficient match kernel (EMK) \cite{bo2010kernel} features are then computed from the spin images, as well as the color features. Next, feature selection is performed to obtain low-dimensional shape and visual descriptors, respectively, using principal component analysis (PCA). These desciptors are combined for subsequent recognition using different classifiers such as linear support vector machine (LinSVM), gaussian kernel support vector machine (kSVM) and random forest. Similar kernel-based descriptors for RGB-D object recognition have also been proposed in \cite{bo2010kernel, bo2011object,bo2011depth,bucak2013multiple}. Covariance matrices-based descriptors that capture visual and shape information have also been explored \cite{tuzel2006region,porikli2006covariance, fehr2016covariance}. Further, approaches that used traditional feature learning techniques such as hierarchical matching pursuit (HMP) \cite{bo2013unsupervised}, PCA \cite{sun2018a}, and single-layer convolutional operators \cite{blum2012learned,cheng2015convolutional} have also been proposed.

\subsection{Learning-based Approaches:}
The remarkable object recognition performance of deep learning-based methods in the case of RGB images has also inspired convolutional neural network (CNN) \cite{gao2019rgb} and transformer-based approaches \cite{tziafas2023early} for RGB-D object recognition.

Among multimodal CNNs for RGB-D object recognition, a common practice is using off-the-shelf CNNs (one per modality) pre-trained on the ImageNet dataset. Typically, the RGB images are directly fed to the corresponding pre-trained network. Depth images, on the other hand, are first rendered as three-channel RGB images to emulate the distribution of the corresponding RGB images before feeding them to the corresponding pre-trained network. Such a rendering is often performed using hand-crafted or learned depth encodings. The most common depth encoding methods include surface normal-based encodings \cite{bo2013unsupervised, aakerberg2017depth}, ColorJet \cite{eitel2015multimodal,schwarz2015rgb}, HHA (stands for the horizontal disparity, the height above ground, and the angle the pixel’s local surface normal makes with the inferred gravity direction) \cite{gupta2014learning}, and embedded depth encoding \cite{zaki2016convolutional}. A learning-based encoding, inspired by methods of colorizing grayscale photographs, is also proposed in \cite{carlucci20182}. Some works employ two or more depth encodings by using multiple depth streams \cite{rahman2017rgb} or ensemble learning \cite{aakerberg2017improving}.

In addition to the choice of depth encodings, such works also focus on appropriately fusing information from the RGB and depth modalities. The two most popular fusion schemes in literature are early and late fusion \cite{sanchez2016comparative}. In the case of early fusion, raw data from both modalities (or their corresponding features) are fused to obtain a joint representation for further joint processing. The central idea of such approaches is to exploit correlations between the two modalities through joint processing. Some works perform this fusion through direct concatenation of fully-connected layers \cite{eitel2015multimodal,aakerberg2017depth, carlucci20182,zhou2019msanet}. In other cases, matrix transformations \cite{wang2015mmss}, canonical correlation analysis (CCA) \cite{wang2015large}, and reconstruction independent component analysis \cite{jin2015partially} have been used for fusion using fully-connected layers. In addition, fusion through a single convolutional layer \cite{liu2018multi} or multiple recurrent neural networks \cite{socher2012convolutional, loghmani2019recurrent, caglayan2022cnns} has also been explored. For instance, Loghmani et al. propose RCFusion \cite{loghmani2019recurrent}, which uses two streams of convolutional networks with the same architecture for multi-level feature extraction from RGB and depth data, followed by a recurrent neural network to fuse the features. On the other hand, late fusion refers to the scenario where most of the feature learning of the two modalities happens independently, and fusion is performed before the last decision stage \cite{zaki2016convolutional,asif2017multi}.  

More recently, approaches that rely on the flexibility of transformers networks to capture cross-modal interactions between multimodal inputs have also been proposed for scene and action recognition \cite{girdhar2022omnivore,girdhar2023omnimae}, semantic segmentation \cite{girdhar2022omnivore,zhang2022cmx}, and object recognition \cite{xiong2023enhancing, tziafas2023early}. For instance, Xiong et al. \cite{xiong2023enhancing} use a pre-trained Vision Transformer (ViT) \cite{dosovitskiy2020image} to extract global features from RGB data and a pre-trained DenseNet to extract local features from RGB and Depth data, which are then fused to feed as input to a classifier for recognition. Alternatively, Tziafas et al. \cite{tziafas2023early} adapt the ViT to perform RGB-D object recognition. They investigate various strategies for depth encoding and fusing the encoded depth with RGB input to show that late fusion of surface normal-based encodings with RGB input works better than the more popularly employed early fusion techniques in the case of RGB-D object recognition.

\section{Mathematical Preliminaries}

The Mapper algorithm \cite{singh2007topological} is an exploratory data analysis tool used to summarize and visualize the structure of any given data. It is based on the idea of using the nerve of a dataset's cover to build a graph (or simplicial complex) representing the data. Specific structures (e.g., loops) can be identified in the graph for subsequent identification of interesting clusters or feature selection.

\begin{figure}
\centering
\includegraphics[width=0.95\textwidth]{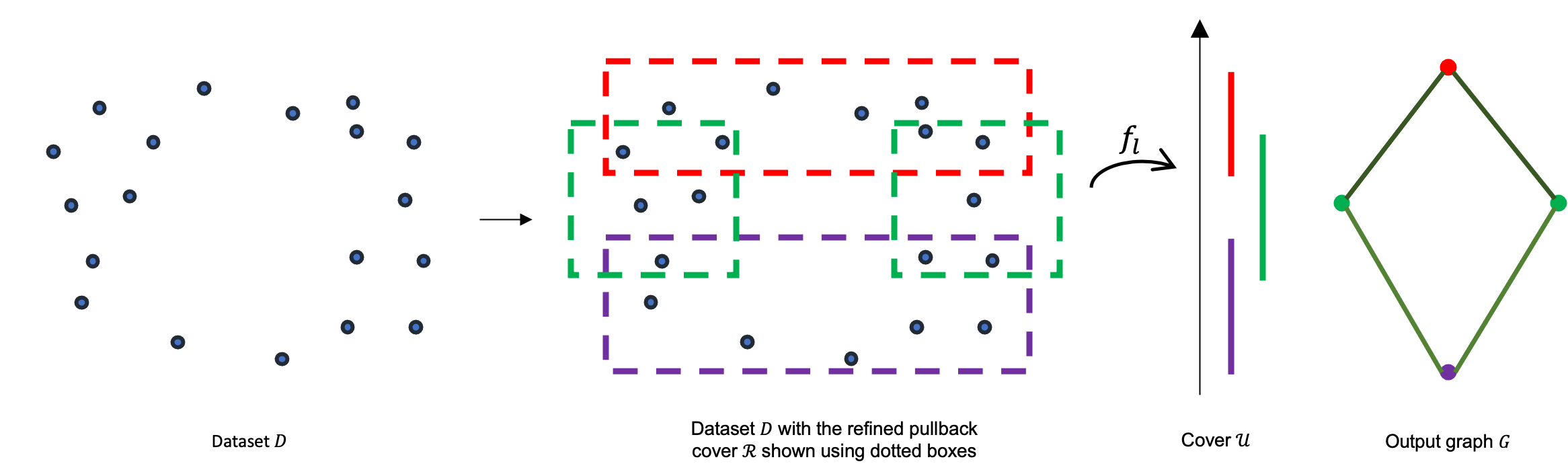}
\caption{An illustration of the Mapper algorithm applied to the sample two-dimensional point cloud. In this example, the height function is used as the lens function, $f_l$, to build the cover in a one-dimensional space. The dotted boxes indicate clusters in the corresponding refined pullback cover of the point cloud. The clusters are collapsed into vertices, and intersections between them are represented by edges in the output graph. Source: modified from \cite{chazal2021introduction}.}
\label{mapperpicture}
\end{figure}

The Mapper algorithm assumes that a high dimensional dataset, say $D$, resides on a low dimensional manifold. Therefore, the first step in the algorithm is to apply a carefully chosen function, $f_l : D \longrightarrow \mathbb{R}$, to reduce the dataset to a low-dimensional space. The choice of the function $f_l$, called the lens function, depends on the features of the data we intend to highlight in the output graph. In the next step, a cover is built in the low-dimensional space. A cover, $U$, of a dataset is defined as a collection of open sets  $\left\{ U_x\right\}_{x \in X}$ ($X$ is the index set) such that every data point is included in at least one $U_x$. Let $\mathcal{U}$ denote the overlapping cover of $f_l (D) $, where $\mathcal{U} = \left\{ \mathcal{U}_y\right\}_{y \in Y}$. The Mapper algorithm then pulls the cover back to $D$ via $f_l^{-1}$ to obtain a collection of subsets $\left\{ f_l^{-1}(\mathcal{U}_y)\right \}_{y \in Y}$ called the \textit{pullback} cover of $D$. Next, a clustering algorithm is applied to each $f_l^{-1}(\mathcal{U}_y)$, where $y \in Y$. The collection of the subsequently obtained clusters\footnote{In case the Mapper algorithm is defined in a topological space, the clustering step splits each $f_l^{-1}(\mathcal{U}_y)$, where $y \in Y$ into connected components, instead of clusters.} is called a \textit{refinement} of $\left\{ f_l^{-1}(\mathcal{U}_y)\right \}_{y \in Y}$. Let $\mathcal{R}$ denote the refined pullback cover of $D$. The output, $G$, of the Mapper algorithm is the nerve of $\mathcal{R}$ \cite{chazal2021introduction}. This nerve is constructed by collapsing each cluster $R \in \mathcal{R}$ into a vertex and creating a $p$-simplex to represent each $(p+1)$-way intersection of $R$'s. Figure \ref{mapperpicture} illustrates the Mapper algorithm applied to a sample point cloud.

\section{Method}
\label{method}
Given an RGB-D image of a cluttered scene, our goal is to recognize all the objects within that scene. We aim to achieve this using only synthetic training data, enabling the applicability of our method for low-cost robots operating in unseen environments. Toward this goal, we propose THOR2, a framework inspired by how object recognition works in humans. We obtain a network of coarse color regions analogous to the MacAdam ellipses to effectively utilize color information within this framework. Section \ref{colornetworkgeneration} describes the generation of the color network, and Section \ref{thor2} describes the THOR2 framework.

\subsection{Color Network Generation}
\label{colornetworkgeneration}





We consider the colors represented by the standard RGB (sRGB) color space, where values for each of the three color channels range from 0 to 255. Let $X_{rgb}$ denote the set of these colors. The sRGB color space is not perceptually uniform; the Euclidean distance between two colors represented using this color space is not proportional to the difference perceived by humans. Therefore, we convert all the elements in $X_{rgb}$ to the CIELAB color space (also known as the $L^\ast a^\ast b^\ast$ color space) to obtain the set $X_{lab}$. We note the CIELAB color space is also not truly perceptually uniform. However, it was purposefully designed such that the Euclidean distance between points is proportional to the perceived color difference. Therefore, it is the preferred color space for algorithmically distinguishing between objects by their color \cite{corke2023robotics}. 


 We then apply the Mapper algorithm to $X_{lab}$ to obtain a color network. Specifically, let $k_{L^\ast}$, $k_{a^\ast}$, and $k_{b^\ast}$ represent the $L^\ast$, $a^\ast$, and $b^\ast$ components of a color $k$ in $X_{lab}$. We use a chroma and hue-based lens function, $f_l$, to transform the three-dimensional data in $X_{lab}$  to a two-dimensional space. We define $f_l$ as follows.

\begin{equation}
    f_l(k) = \left (\sqrt{k_{a^\ast}^2+k_{b^\ast}^2} , \xi + \arctan \left(\frac{k_{b^\ast}}{k_{a^\ast}}\right) \right ),
\end{equation}

\noindent where $\xi$ is a constant offset selected based on the cover. We adopt the standard choice \cite{chazal2021introduction} of building a cubical cover $\mathcal{U}$ by considering a set of regularly spaced intervals of equal length covering the set $f_l(X_{lab})$. Let $r_1$ and $r_2$ (where $r_1, r_2>0$) denote the lengths of the intervals (also known as the \textit{resolution} of the cover) along the two dimensions of $f_l(X_{lab})$, respectively. Let $g_1$ and $g_2$ denote the respective percentages of overlap (also known as the \textit{gain} of the cover) between two consecutive intervals. Subsequently, the Mapper algorithm applies a clustering algorithm to $f_l^{-1}(U)$ for each $U \in \mathcal{U}$ to obtain the refined pullback cover $\mathcal{R}$ of $X_{lab}$. For clustering, we use the HyAB distance metric \cite{abasi2020distance} for computing the distance between two colors. Unlike other color difference formulae (e.g., CIEDE2000 \cite{luo2001development}), the HyAB distance is applicable to a large range of color differences required for practical applications \cite{abasi2020distance}. The HyAB distance between two colors $m$ and $n$ in the CIELAB space is defined as

\begin{equation}
\hyab(m,n) = \lvert m_{L^\ast} - n_{L^\ast} \rvert + \sqrt{(m_{a^\ast} - n_{a^\ast})^2 + (m_{b^\ast} - n_{b^\ast})^2 }.
\end{equation}

\begin{figure}
\centering
\includegraphics[width=\textwidth]{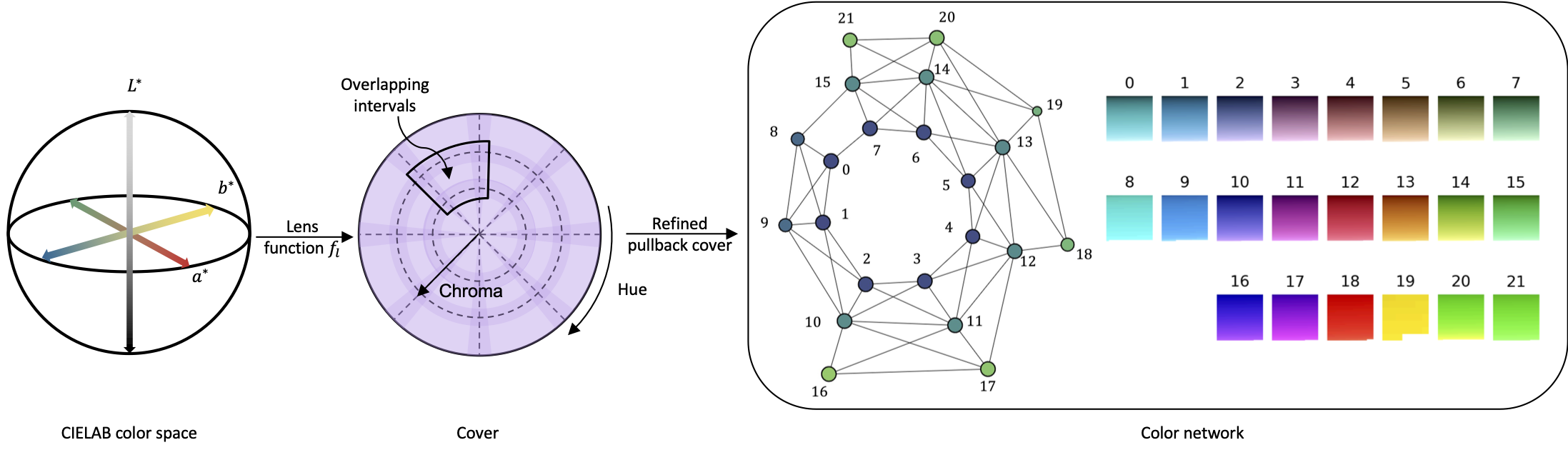}
\caption{Generation of a network of coarse color regions using the Mapper algorithm. A chroma and hue-based lens function projects colors from the CIELAB space to a two-dimensional space. Clustering is performed in each of the overlapping intervals of the cover to obtain a refined pullback cover whose nerve represents the color network.}
\label{mapperoutput}
\end{figure}

\noindent Next, the nerve of the refined pullback cover is obtained to get the output of the Mapper algorithm. Note that the cyclic nature of the chroma-related dimension of $f_l(X_{lab})$ cannot be captured through the cubical cover described above. Therefore, we augment the output of the Mapper algorithm to add edges connecting the nodes corresponding to the first and last intervals along that dimension. Subsequently, we eliminate redundant nodes, as described in \ref{colornetworkimpl}, to obtain the final color network shown in Figure \ref{mapperoutput}. In this network, the nodes represent useful color regions identified by the Mapper algorithm, and the edges represent the overlap between the corresponding color regions.

Let $G=(V,E)$ be the color network, representing a non-empty set of vertices or nodes $V$ and a set of edges $E$. Let $n_c$ represent the numbers of vertices in $G$, i.e., the number of color regions. We define a similarity matrix, $\Delta$, of size $n_c \times n_c$ to capture the similarity and connectivity between the different color regions in $G$. We define $\Delta$ as follows

\begin{equation}
\Delta = 
\begin{bmatrix}
    \delta_{11} & \delta_{12} &  \dots  & \delta_{1n_c} \\
    \delta_{21} & \delta_{22} &  \dots  & \delta_{2n_c} \\
    \vdots & \vdots & \vdots &  \vdots \\
    \delta_{n_c1} & \delta_{n_c2} & \dots  & \delta_{n_cn_c}
\end{bmatrix}    
\end{equation}
\noindent where $\delta_{i^\prime j^\prime}$ represents the similarity between the $i^\prime$-th and $j^\prime$-th nodes. 
Since every edge in $E$ does not represent the same perceptual difference between the color regions, first, we assign a weight to each edge. Let $\eta_{i^\prime j^\prime}$ be an edge in $E$ that connects the $i^\prime$-th and $j^\prime$-th nodes. To assign the weight, first, we compute the mean color\footnote{To compute the mean, we convert all the colors in that node to the sRGB space, calculate the arithmetic mean of the intensities in each channel \cite{kasaei2021investigating}, and convert the resulting mean color back to the CIELAB space} of the $i^\prime$-th and $j^\prime$-th colors nodes. The edge $\eta_{i^\prime j^\prime}$ is then assigned a weight equal to the HyAB distance between the mean colors of the $i^\prime$-th and $j^\prime$-th nodes. We then set $\delta_{i^\prime j^\prime} = \frac{1}{1 + l_{i^\prime j^\prime}}$, where $l_{i^\prime j^\prime}$ is the weight of minimum weight path connecting the $i^\prime$-th and $j^\prime$-th nodes. This similarity matrix is pre-computed and used in the TOPS2 descriptor computation for THOR2, as described in the following section.

\subsection{THOR2}
\label{thor2}

Similar to THOR, THOR2 takes inspiration from human object recognition. According to cognitive neuropsychology \cite{rapp2015handbook,ward2015student}, humans first process the basic components of an object, such as color, depth, and form. Subsequently, they group these components to segregate surfaces into figure and ground (analogous to foreground segmentation). The object is then rotated to standard orientation to obtain a `normalized' view, i.e., to achieve object constancy, which is the ability to understand that objects remain the same irrespective of viewing conditions. Next, structural descriptions of the normalized view are computed and matched with those stored in memory. Last, these representations are associated with semantic attributes for recognition.

\begin{figure} 
\begin{subfigure}{\columnwidth}
  \centering
  \includegraphics[width=\textwidth]{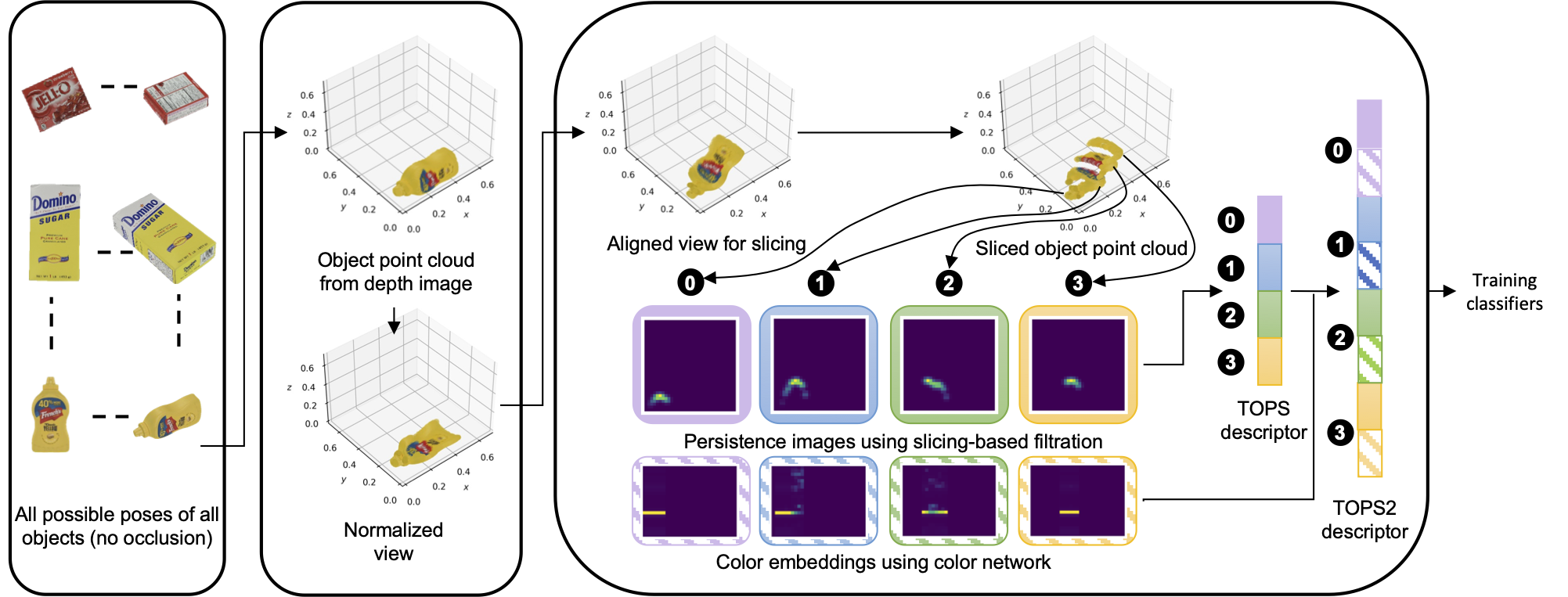} 
  \caption{Training stage for THOR2 showing a visualization of the TOPS and TOPS2 descriptors computed from the synthetic RGB-D image corresponding to a sample object pose (of an unoccluded mustard bottle) considered during training.}
  \label{training}
\end{subfigure}

\begin{subfigure}{\columnwidth}
  \centering
  \includegraphics[width=\textwidth]{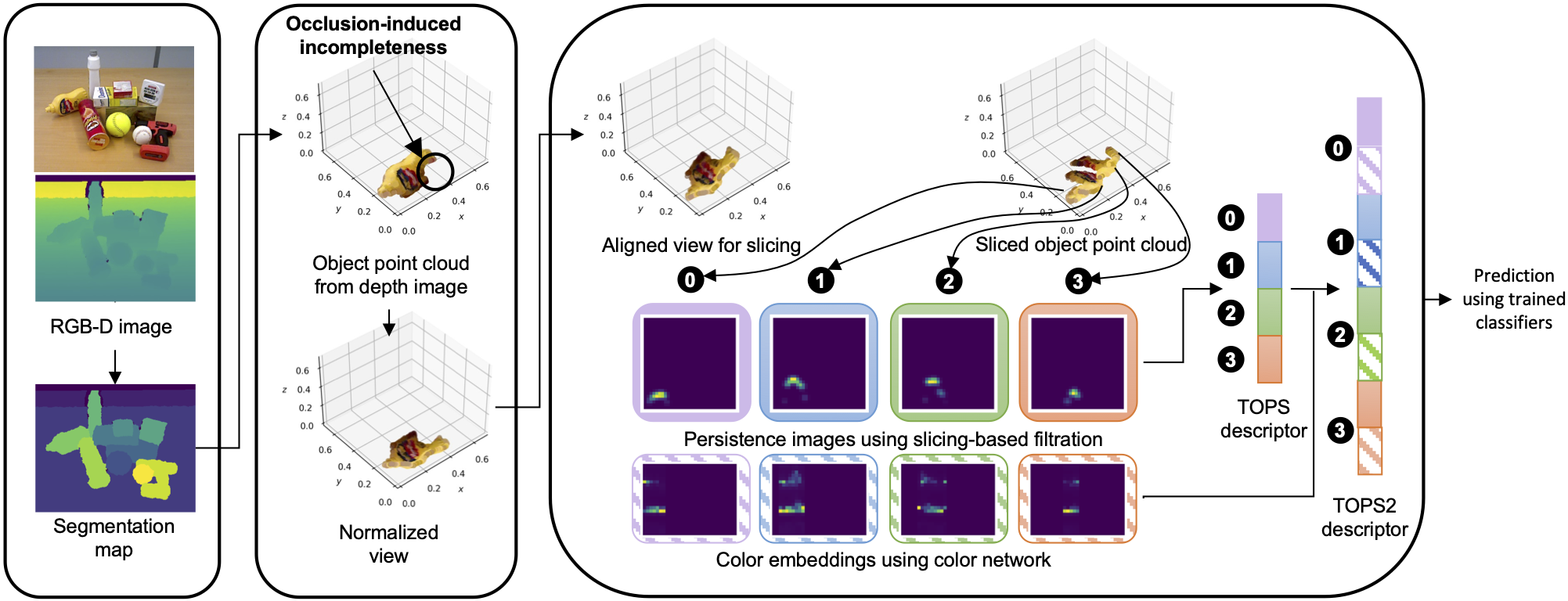}  
  \caption{Testing stage for THOR2, aligned with human object recognition stages, shows a visualization of an occluded object's TOPS and TOPS2 descriptors computed from a real RGB-D scene image.}
  \label{testing}
\end{subfigure}

\caption{Proposed framework, THOR2, for 3D shape and color-based recognition using object unity, facilitated by the similarity in the TOPS and TOPS2 descriptors of unoccluded and occluded objects. In this illustration, the persistence images and color embeddings for the first three slices of the mustard bottle denoted using the purple, blue, and green boxes are similar across the training and test stages; only the persistence image and color embedding corresponding to the last slice of the mustard bottle (which is affected by occlusion) are different.}
\label{pipeline}
\end{figure}

In the case of THOR2, first, we obtain colored point clouds for every object in the RGB-D image of the scene with the help of instance segmentation maps. As this work focuses on object recognition, we assume these maps are available using methods such as those in \cite{xie2021unseen, lu2023self}. In the second stage, we perform view-normalization on these point clouds, as described in Section \ref{viewnormalization}. In the third stage, we compute the TOPS and TOPS2 descriptors of the view-normalized point clouds, as described in Section \ref{descriptorcomputation}. In the last stage, these descriptors are appropriately fed to classifiers for recognition. Section \ref{trainingtesting} describes this `testing' of classifiers and their training using synthetic data. Fig \ref{pipeline} depicts the overall THOR2 framework.



\subsubsection{View Normalization.} 
\label{viewnormalization}
Consider a colored object point cloud $\mathcal{P}$ in $\mathbb{R}^{3}$. First, we use an approximation of the O'Rourke's algorithm \cite{o1985finding} to compute the minimal volume bounding box of $\mathcal{P}$. We orient this bounding box such that the coordinate axes are ordered with respect to the principal components. Subsequently, the point cloud is rotated so the computed bounding box aligns with the coordinates axes. We fine-tune this approximate alignment as follows. Specifically, the point cloud is further rotated such that the 2D bounding boxes of its projection on the $x-y$ and $y-z$ planes are aligned with their respective coordinate axes. Last, we perform translation so that the resultant view-normalized point cloud, $\tilde{\mathcal{P}}$, lies in the first octant.

\subsubsection{TOPS and TOPS2 Descriptor Computation}
\label{descriptorcomputation}

The TOPS descriptor captures the shape of a point cloud through the persistence images of its slices. These persistence images are generated by employing a slicing-based approach to compute topological features using persistent homology \cite{samani2024persistent}. The TOPS2 descriptor, an extension of the TOPS descriptor, captures a point cloud's color and 3D shape through interleaved color embeddings and persistence images, respectively.

\threesubsection{\textbf{TOPS descriptor computation}} In \cite{samani2024persistent}, the TOPS descriptor is computed as follows. Consider a view-normalized point cloud, $\tilde{\mathcal{P}}$. First, it is rotated about the $y$-axis such that its longitudinal axis makes an angle $\alpha$ with the $x-y$ plane to point cloud as $\hat{\mathcal{P}}$. Then, $\hat{\mathcal{P}}$ is sliced along the $z$-axis. Effectively, this rotation of $\tilde{\mathcal{P}}$ determines the relative direction of its slicing. Then, $\hat{\mathcal{P}}$ is sliced along the $z$-axis to get slices $\mathcal{S}^i$, where $i \in \mathbb{Z} \cap [0,\frac{h}{\sigma_1}]$. Here, $h$ is the dimension of the axis-aligned bounding box of $\hat{\mathcal{P}}$ along the $z$-axis, and $\sigma_1$ is the thickness of the slices. Let $p=(p_x,p_y,p_z)$ be a point in $\hat{\mathcal{P}}$. The slices $\mathcal{S}^i$ are then obtained as follows. 
\begin{equation}
\mathcal{S}^{i}:=\left\{p \in \hat{\mathcal{P}} \mid i\sigma_1 \leq p_z < (i+1)\sigma_1    \right\}.
\label{firstslicing}
\end{equation}
For every slice, $\mathcal{S}^i$, a filtration of simplicial complexes that mimics further slicing of the slice along the $x$-axis is constructed. Persistent homology is applied to the filtration to obtain a persistence image, $\mathcal{I}^i$, that captures the detailed shape of the slice. Last, the persistence images are vectorized and stacked to obtain the TOPS descriptor. We refer the reader to the work in \cite{samani2024persistent} for additional details on the computation of the persistence images. 

\threesubsection{\textbf{TOPS2 descriptor computation}}
The TOPS2 descriptor of a point cloud is computed as follows. Let $s=(s_x,s_y,s_z)$ represent a point in a slice, $\mathcal{S}^i$, of $\hat{\mathcal{P}}$ obtained according to Equation \ref{firstslicing}. For every slice $\mathcal{S}^i$, we modify the $z$-coordinates $\forall s \in \mathcal{S}^i$ to $s_z^\prime$, where $s_z^\prime=i\sigma_1$. Color embeddings for every slice are then obtained as follows

Consider a slice $\mathcal{S}^i$. We perform further slicing of $\mathcal{S}^i$ along the $x$-axis to obtain \textit{strips} $\Omega^j$, where $j \in \mathbb{Z} \cap [0,\frac{w}{\sigma_2}]$. Here, $w$ is the dimension of the axis-aligned bounding box of the slice along the $x$-axis and $\sigma_2$ represents the 'thickness' of a strip. For every strip $\Omega^j$, we obtain corresponding color vectors $\Phi^j = \big[\begin{smallmatrix} \phi_{1} & \phi_{2} &  \dots  & \phi_{n_c} \end{smallmatrix}\big]^T $ as follows.

\begin{equation}
\phi_\lambda = \sum_{\omega  \: \in \:  \Omega^{j}} \frac{\mathbbm{1}_{X^\lambda_{rgb}}(\omega)}{\sum\limits_{\lambda = 1}^{n_c}\mathbbm{1}_{X^\lambda_{rgb}}(\omega)},
\end{equation}

\noindent where $\lambda \in \{1, \ldots, n_c\}$ represents the $\lambda$-th color region, $\omega$ represents the color of a point in $\Omega^j$ (in the sRGB color space), $\mathbbm{1}$ denotes the indicator function of a set, and $X^\lambda_{rgb}$ represents the set of colors (in the sRGB color space) belonging to the $\lambda$-th color region. Consequently, the color vectors $\Phi^j$ approximately represent the color constitution (in terms of the color regions) of the strips $\Omega^j$.

We then stack the color vectors (with appropriate zero padding) to obtain an $n_s^{max} \times n_c$ dimensional color matrix $\mathcal{C}^i$. Let $\mathcal{C}^i = \big[\begin{smallmatrix}  \mathbf{O}& \ldots & \Phi_{1} & \Phi_{2} & \ldots & \Phi_{n_s} & \ldots & \mathbf{O} \end{smallmatrix}\big]^T $, where $n_s$ is the number of strips in the corresponding slice $\mathcal{S}^i$, $n_s^{max}$ is the maximum number of strips in any given slice, and $\mathbf{O}$  represents a $ n_c \times 1$ dimensional zero matrix. Consequently, the color color matrix  $\mathcal{C}^i$ approximately represents the color constitution (in terms of the color regions) of the slice $\mathcal{S}^i$ in a spatially-aware manner. Last, we obtain an embedding $\mathcal{E}^i$ corresponding to the color matrix $\mathcal{C}^i$ as follows.

\begin{equation}
    \mathcal{E}^i = (\mathcal{C}^i\Delta)^T
\end{equation}

\begin{figure*}
\centering
\includegraphics[width=\textwidth]{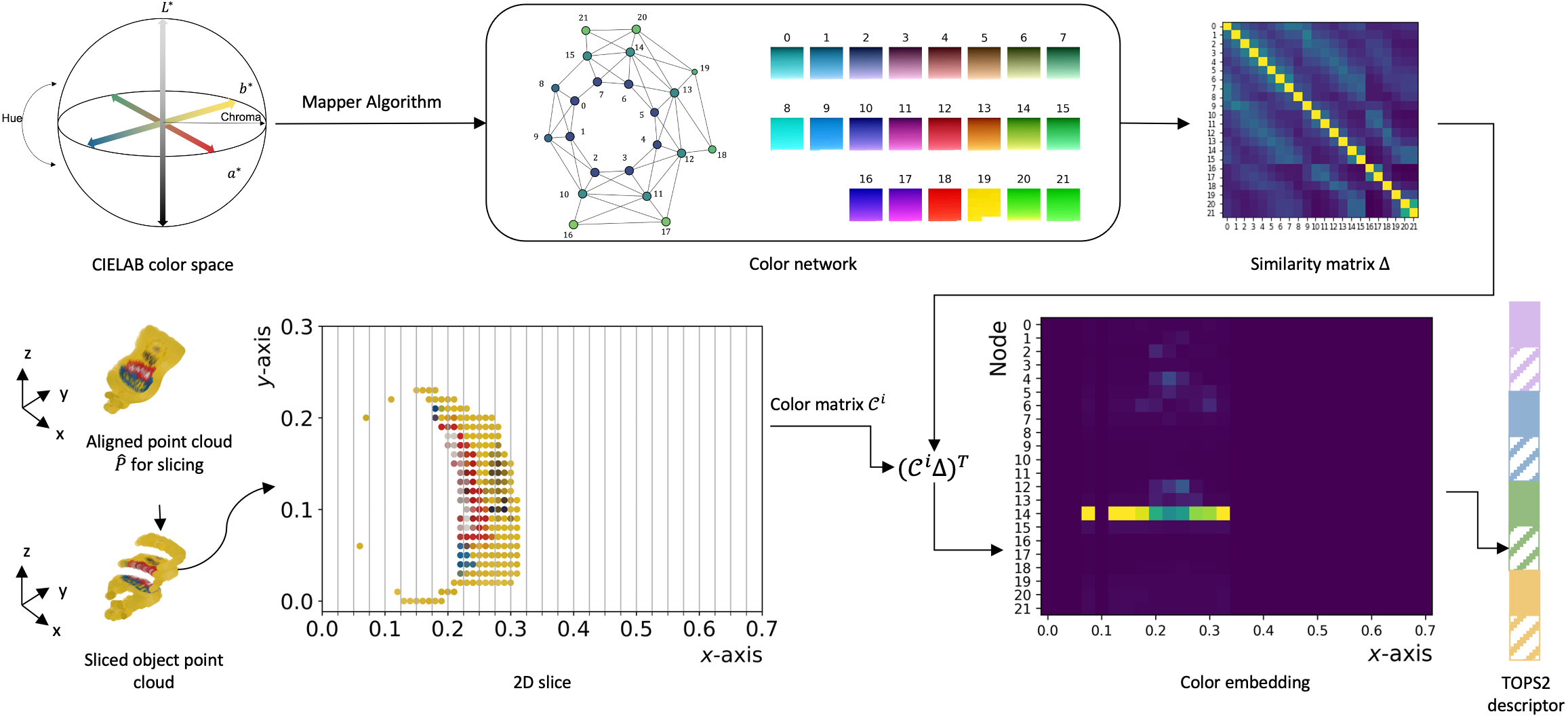}
\caption{Computation of the TOPS2 descriptor from the slices of an aligned object point cloud, $\hat{\mathcal{P}}$. The similarity matrix, $\Delta$, representing the color network obtained using the Mapper algorithm, is used to compute color embeddings for every slice of the point cloud. The resulting color embeddings are vectorized (shown using patterned rectangles) and interleaved with the vectorized persistence images (solid rectangles) to obtain the TOPS2 descriptor.}
\label{embeddingfigure}
\end{figure*}

We then vectorize and concatenate both the persistence image, $\mathcal{I}^i$, and the embedding, $\mathcal{E}^i$, to obtain a combined shape and color-based representation for $\mathcal{S}^i$. Such combined representations of all the slices are stacked to obtain the TOPS2 descriptor. Figure \ref{embeddingfigure} shows the computation of the TOPS2 descriptor for a sample object.

\subsubsection{Training and Testing of Classifiers}
\label{trainingtesting}

\threesubsection{\textbf{Training}}
Point clouds generated from RGB-D images are often incomplete, with the degree of incompleteness varying based on the camera's orientation relative to the object. Therefore, we consider synthetic RGB-D images corresponding to all the possible views of all the objects. As with THOR, we do not consider object occlusion scenarios in our training set. We begin by generating colored object point clouds from the RGB-D images and scale them by a factor of $\sigma_s$. Next, we perform view normalization and simultaneously compute the TOPS and TOPS2 descriptors for the point clouds. We train one classifier, a multi-layer perceptron, $M_{1}$, using the TOPS descriptor and another multi-layer perceptron, $M_{2}$, using the TOPS2 descriptor. 

\threesubsection{\textbf{Testing}}
When testing, i.e., recognizing objects from a real RGB-D image of a cluttered scene, first, we generate the individual colored point clouds of all the objects in the scene using the instance segmentation maps. Similar to the training stage, we scale the point clouds by a factor of $\sigma_s$. Consider a scaled object point cloud $\mathcal{P}_t$ obtained from the test image. To recognize $\mathcal{P}_t$, first, we perform view normalization to obtain $\tilde{\mathcal{P}_t}$. Similar to THOR, we then determine if the object corresponding to $\tilde{\mathcal{P}_t}$ is occluded. To do so, first, we extract the object's contour from the segmentation map. For each pixel along the contour, we classify it as part of an occlusion boundary if at least one neighboring pixel is labeled as an object instance and has a depth value lower than that pixel's own depth value. If the object is occluded, we rotate $\tilde{\mathcal{P}_t}$ by $\pi$ about the $z$-axis to ensure that the first slice on the occluded end of the object is not the first slice during subsequent TOPS and TOPS2 descriptor computation. We then compute the TOPS and TOPS2 descriptors corresponding to $\tilde{\mathcal{P}_t}$ and use the corresponding classifier models, $M_1$ and $M_2$, respectively, to obtain two predictions. We choose the prediction with the highest probability as our final prediction.



\section{Experiments}

In this section, we describe the experimental evaluation of THOR2. Two benchmark datasets, described in Section \ref{datasets}, are used for the evaluation. We compare the performance of THOR2 with the performances of the shape-based THOR framework \cite{samani2024persistent} and three other deep learning-based end-to-end models for RGB-D object recognition; namely, RCFusion \cite{loghmani2019recurrent}, ViT-CNN \cite{xiong2023enhancing}, and RGB-D ViT \cite{tziafas2023early}. Section \ref{implementationdetails} describes the implementation details, and the results from these evaluations are reported in Section \ref{results}. We also implement THOR2 on a real-world robot as described in Section \ref{robotimpl}
 
\subsection{Datasets}
\label{datasets}
We evaluate the performance of our method on two benchmark datasets: the YCB10 subset of the OCID dataset \cite{suchi2019easylabel} and the UW-IS Occluded dataset \cite{samani2024persistent}.

\subsubsection{OCID Dataset}

The YCB10 subset of the OCID dataset comprises 24 image sequences featuring increasingly cluttered scenes containing up to ten objects each. The sequences are categorized into three types: cuboidal (objects with sharp edges), curved (objects with smooth curved surfaces), and mixed (a combination of cuboidal and curved objects). Each sequence is captured using two RGB-D cameras, named the `lower camera' and `upper camera,' positioned at different heights and angles to replicate typical robotic system configurations. Additionally, the dataset provides temporally smoothed, point-wise labeled point clouds for every frame within the sequence.

\subsubsection{UW-IS Occluded Dataset}
The UW-IS Occluded dataset includes two distinct indoor environments: a lounge where the objects from the YCB \cite{calli2015benchmarking} and BigBird \cite{singh2014bigbird} datasets are positioned on a tabletop, and a mock warehouse where objects are arranged on shelves. Each environment provides RGB-D images from 36 videos, capturing five to seven objects per scene, shot from distances up to approximately 2 meters using an Intel RealSense D435 camera. The videos systematically cover varying levels of object occlusion across three categories of objects and are recorded under two different lighting conditions.

\subsection{Implementation Details}
\label{implementationdetails}

We provide implementation details associated with the generation of the color network in Section \ref{colornetworkimpl}, THOR2 in Section \ref{thor2impl}, and all the other comparison methods in Section \ref{comparisonmethodsimpl}

\subsubsection{Color Network Generation}
\label{colornetworkimpl}
We use the Kepler Mapper library \cite{KeplerMapper_JOSS, KeplerMapper_v1.4.1-Zenodo} for applying the Mapper algorithm to $X_{rgb}$. $X_{rgb}$ consists of the colors obtained by uniformly sampling each color channel with a step size of five to obtain a total of $52^3$ color values. We use the CIE standard illuminant D65 (and its color space chromaticity coordinates corresponding to the standard $2\degree$ observer) to convert colors from the sRGB space to the CIELAB color space. Since the Mapper algorithm is an exploratory data analysis tool, a commonly used strategy is to explore a range of associated parameters and select those that provide the most informative output (with respect to the user perspective) \cite{chazal2021introduction}. We choose $\xi = \frac{\pi}{8}$, $g_1 = 10\%$, and $g_2=25\%$ for our purposes. We set $r_1$ and $r_2$ to divide the corresponding dimensions into three and eight equally-spaced intervals, respectively. We use the DBSCAN algorithm \cite{ester1996density} with HyAB as the distance metric for clustering. During clustering, a point is considered a core point if it has at least six data points in its neighborhood (points that are at most seven units apart are considered neighbors). Once the Mapper output is obtained, we perform post-processing to eliminate the redundant nodes. First, we identify the node pairs that have more than $95\%$ members in common. For each pair, we then compute the mean colors of the nodes. If the mean colors are neighbors, we appropriately merge the nodes to form a single node.

\subsubsection{THOR2}
\label{thor2impl}
The details associated with the TOPS and TOPS2 descriptor computation, synthetic training data generation, and the training and testing of classifiers are described below.

\threesubsection{\textbf{Descriptor computation}} We compute the TOPS descriptor as described in \cite{samani2024persistent}. We follow their parameter choices and choose the scale factor $\sigma_s = 2.5$ and set $\sigma_1 = 0.1$, $\sigma_2 = 2.5 \times 10^{-2}$, and $\alpha = \frac{\pi}{4}$ to compute the suitable TOPS2 descriptors. Consequently, $n_s^{max}$ is $29$ and $31$ for the OCID and UW-IS Occluded datasets, respectively. 

\threesubsection{\textbf{Synthetic training data generation}} \label{syndatageneration}We consider objects appearing in the OCID and UW-IS Occluded datasets for which the scanned meshes are available from \cite{calli2015benchmarking}. We obtain synthetic RGB-D images of those objects using the Panda3D \cite{panda3d_2018} framework as follows. We position each object at the center of an imaginary sphere and generate RGB-D images considering different camera positions on the sphere. Camera positions are obtained by considering azimuthal angles ranging from $[0,2\pi)$ in increments of $\frac{\pi}{36}$ and polar angles ranging from $[0,\pi]$ in increments of $\frac{\pi}{36}$. The depth images are produced with a scale of 0.001 m, meaning each unit increase in depth corresponds to a 1 mm increment in simulation.

\threesubsection{\textbf{Training and Testing of Classifiers}}
We use workstations with GeForce GTX 1080 and 1080 Ti GPUs running Ubuntu 18.04 LTS for training (and testing) the classifier models. Separate TOPS-based and TOPS2-based classifiers are trained for every sequence using synthetic RGB-D images in the case of the OCID dataset. In the case of the UW-IS Occluded dataset, we train a single TOPS-based classifier and a single TOPS2-based classifier. For training, first, we augment the training set by mirroring all the view-normalized point clouds across the $x$ and $y$ coordinate axes (in place). The classifiers are five-layer fully connected networks, i.e., Multi-layer Perceptrons (MLPs) \revision{with layers having 512, 256, 128, 64, and} \texttt{n\char`_classes} \revision{nodes, respectively, where} \texttt{n\char`_classes} \revision{represents the number of object classes}. During training, we use the Adam optimizer to optimize the categorical cross-entropy loss over 100 epochs. The learning rate is set to $10^{-2}$ for the first 50 epochs and is decreased to $10^{-3}$ for the subsequent 50 epochs. 

For testing on the OCID dataset, we use the temporally smoothed, point-wise labeled point clouds provided for every frame in the sequence. The UW-IS Occluded dataset is recorded using commodity hardware and consists of noisy depth images. \revision{Instead of existing approaches for denoising depth images} \cite{yan2020depth}, we consistently perform pre-processing via voxel grid-based downsampling and outlier removal for every object point cloud, as described in \cite{samani2024persistent}. 
We report the  test results from five-fold cross-validation for THOR2 (and all the comparison methods).

\subsubsection{Comparison methods}
\label{comparisonmethodsimpl}
We use the official implementations available from \cite{samani2024persistent} and \cite{loghmani2019recurrent} to evaluate THOR and RCFusion, respectively. In the case of ViT-CNN \cite{xiong2023enhancing}, we use the VitMAE model available from the transformers library \cite{wolf2020transformers} to perform RGB feature extraction. The DenseNet-161 model available from the PyTorch Hub is used to extract the CNN-based RGB-D features. Owing to the large size of features and the large training dataset, we train a multi-layer perceptron in this case instead of the proposed kNN $(k=1)$ nearest neighbor classifier. In the case of RGB-D ViT \cite{tziafas2023early}, we use the base configuration for ViT initialized using weights from pretraining on ImageNet (provided by PyTorch). We train all the methods using the synthetic data as described in Section \ref{syndatageneration}. 

Surface normal representations of the depth images for RCFusion, ViT-CNN, and RGB-D ViT are obtained according to \cite{caglayan2022cnns}. During the training of ViT-CNN and RGB-D ViT, we use the mean and standard deviation corresponding to the ImageNet dataset to standardize the normalized RGB images and surface normal representations of the synthetic dataset. For testing both the models, first, depth interpolation is performed \cite{caglayan2022cnns}, and object crops are obtained using the instance segmentation maps. Subsequently, we replace the background (i.e., background pixels of the RGB image and the corresponding values of the surface normal representation of the depth image) to match that of the synthetic images used for training. The object crops are then resized and padded appropriately to achieve the required input size (i.e., $224 \times 224$). Note that we perform the same procedure for background replacement, resizing, and padding RGB-D images from the YCB dataset in the experiments where those real-world images are used to train the \revision{RCFusion, ViT-CNN, and RGB-D ViT models}.

\subsection{Results}
\label{results}
In the first stage of evaluation, we compare the performance of the THOR2 framework with the other end-to-end object recognition methods. Section \ref{resultsonendtoendmodels} summarizes our findings from this evaluation. Next, in Section \ref{uwis2ablationresults}, we report observations from ablation studies to analyze the components of the THOR2 framework. In Section \ref{realworldtrainingexps}, we evaluate the performance of one of the comparison methods (RGB-D ViT) when varying amounts of real-world data are used to train the model in addition to the synthetic training data.

\subsubsection{Comparison with End-to-end Methods}
\label{resultsonendtoendmodels}

We compare the performance of THOR2 with THOR (with the MLP library), which only uses 3D shape information captured by the TOPS descriptor for recognition. In addition, we also compare THOR2's performance with other end-to-end models for RGB-D object recognition; namely, RCFusion, ViT-CNN, and RGB-D ViT. Overall, we note that THOR2 achieves sizeable performance improvements over THOR with the help of additional color information while outperforming all the end-to-end models on both the OCID and UW-IS Occluded datasets.

\begin{figure*}
    \centering
    \includegraphics[width=\textwidth]{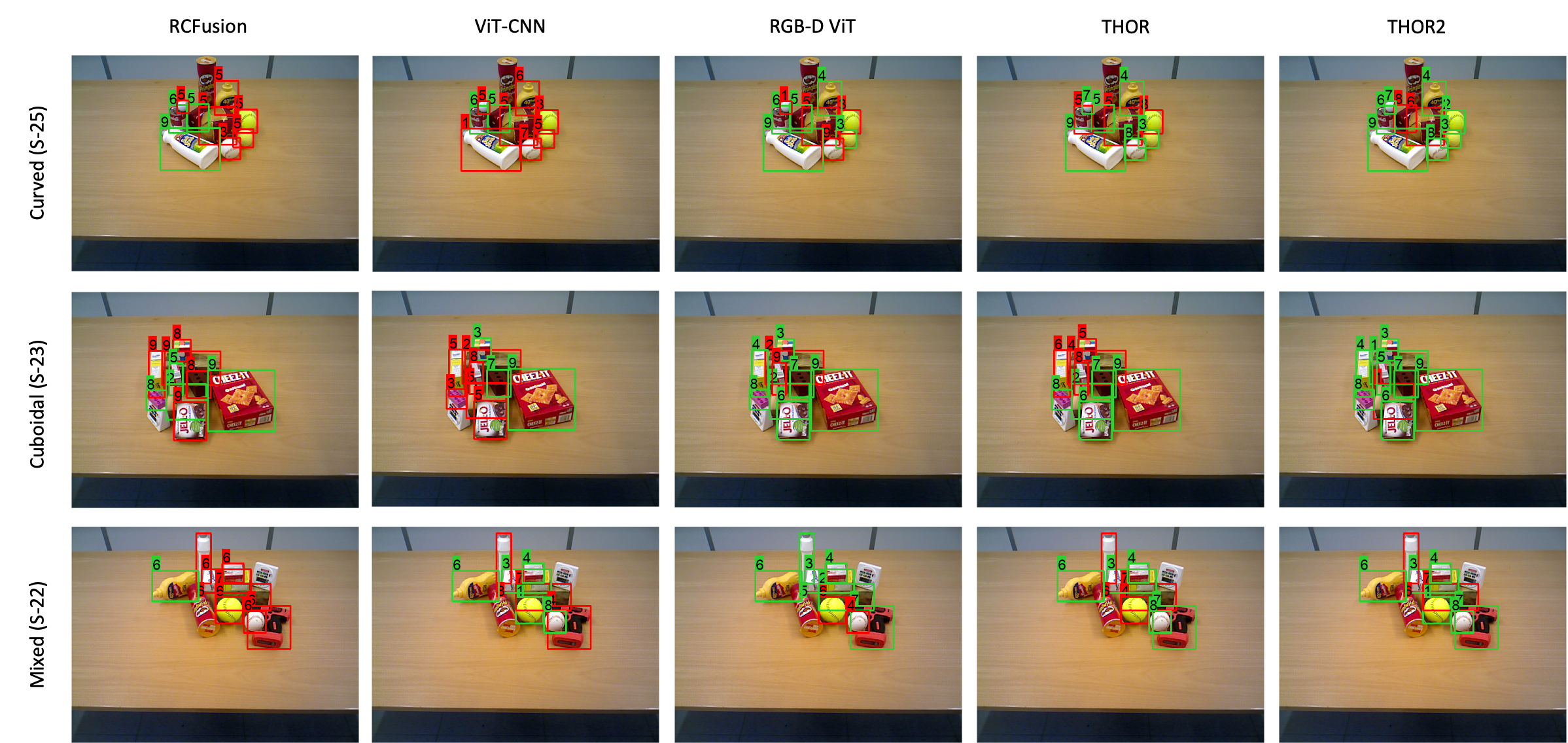}
    \caption{Sample results from the OCID dataset sequences recorded using the lower camera (green and red boxes represent correct and incorrect recognition, respectively). The three rows show results on sequences with curved, cuboidal, and mixed objects placed on a table.}.
    
    \label{ocidresultsamples}
\end{figure*}

\threesubsection{\textbf{Evaluation on the OCID dataset}}
Table \ref{synocidlower} reports the performance of all the methods on the test sequences of the OCID dataset recorded using the lower camera. Overall, THOR2 achieves approximately $8.5\%$ higher recognition accuracy than THOR and approximately $18.9\%$ higher accuracy than RGB-D ViT, which is the best-performing model among all the end-to-end models (i.e., RCFusion, ViT-CNN, and RGB-D ViT). We note that incorporating color information in THOR2 is particularly beneficial in the case of curved and cuboidal object sequences where object geometries are similar and less easily distinguishable. For instance, the color information in THOR2 helps distinguish between a softball and a baseball in S-25 (see the first row in Figure \ref{ocidresultsamples}). In another instance, color helps distinguish between the sugar box and the pudding box in S-23 (see the second row in Figure \ref{ocidresultsamples}). THOR2 achieves substantially better performance than THOR in such sequences. Performance improvements are also observed for all but two mixed object sequences where the object geometries vary considerably. We refer the reader to Section \ref{instancesegerrors} for a discussion on the performance of THOR and THOR2 in those two sequences (i.e., S-02 and S-07). Furthermore, we note that among all the methods that use both shape and color information, THOR2 is the best-performing method in all but one sequence (i.e., S-36, which is discussed in Section \ref{specificocclusionscenarios}). 


\begin{table}[]
\centering
\caption{Comparison of mean recognition accuracy (in \%) of THOR2 with end-to-end models on the OCID dataset sequences recorded using the lower camera (best in bold)}
\label{synocidlower}
\resizebox{\textwidth}{!}{%
\begin{tabular}{@{}ccc|ccccc@{}}
\toprule
Place                   & Scene type              & Seq. ID & RCFusion         & ViT-CNN                    & RGB-D ViT                 & THOR                      & THOR2                     \\ \midrule
\multirow{12}{*}{Table} & \multirow{4}{*}{Curved} & S-25  & 38.28 $\pm$ 3.13 & 46.59 $\pm$ 2.24          & 49.10 $\pm$ 2.64          & 65.31 $\pm$ 2.03          & \textbf{74.90 $\pm$ 4.51} \\
                        &                         & S-26  & 30.30 $\pm$ 3.87 & 37.56 $\pm$ 3.87          & 64.06 $\pm$ 2.82          & 57.18 $\pm$ 2.25          & \textbf{75.70 $\pm$ 2.71} \\
                        &                         & S-35  & 31.28 $\pm$ 4.12 & 29.06 $\pm$ 2.94          & 34.59 $\pm$ 2.95          & 49.83 $\pm$ 2.07          & \textbf{64.38 $\pm$ 3.83} \\
                        &                         & S-36  & 26.17 $\pm$ 5.29 & 31.05 $\pm$ 2.07          & \textbf{67.91 $\pm$ 1.95} & 61.35 $\pm$ 0.49          & 63.33 $\pm$ 0.98          \\
                        \cmidrule(l){2-8} 
                        & \multirow{4}{*}{Cuboid} & S-23  & 51.53 $\pm$ 5.25 & 43.10 $\pm$ 1.93          & 62.88 $\pm$ 5.29          & 72.82 $\pm$ 0.94          & \textbf{83.75 $\pm$ 2.35} \\
                        &                         & S-24  & 61.59 $\pm$ 3.26 & 37.97 $\pm$ 3.93          & 58.25 $\pm$ 2.49          & 57.01 $\pm$ 4.28          & \textbf{70.53 $\pm$ 1.60} \\
                        &                         & S-33  & 54.50 $\pm$ 6.55 & 46.45 $\pm$ 2.87          & 55.38 $\pm$ 2.56          & 65.90 $\pm$ 3.27          & \textbf{77.54 $\pm$ 3.47} \\
                        &                         & S-34  & 41.98 $\pm$ 8.15 & 52.84 $\pm$ 2.94          & 53.98 $\pm$ 3.63          & 65.09 $\pm$ 3.38          & \textbf{79.86 $\pm$ 1.28} \\
                        \cmidrule(l){2-8} 
                        & \multirow{4}{*}{Mixed}  & S-21  & 51.53 $\pm$ 3.08 & 49.50 $\pm$ 4.51          & 48.27 $\pm$ 2.06          & \textbf{68.02 $\pm$ 1.06} & \textbf{71.49 $\pm$ 3.85} \\
                        &                         & S-22  & 32.12 $\pm$ 5.56 & 40.46 $\pm$ 2.32          & \textbf{80.22 $\pm$ 4.20} & 62.11 $\pm$ 3.77          & \textbf{73.68 $\pm$ 2.43} \\
                        &                         & S-31  & 43.89 $\pm$ 3.44 & 36.03 $\pm$ 2.12          & \textbf{81.07 $\pm$ 3.76} & \textbf{74.59 $\pm$ 1.34} & \textbf{79.95 $\pm$ 4.58} \\
                        &                         & S-32  & 38.11 $\pm$ 1.83 & \textbf{70.93 $\pm$ 1.64} & 58.93 $\pm$ 3.80          & 67.16 $\pm$ 1.76          & \textbf{74.25 $\pm$ 1.87} \\
\midrule
\multirow{12}{*}{Floor} & \multirow{4}{*}{Curved} & S-05  & 12.06 $\pm$ 0.95 & 48.26 $\pm$ 1.96          & 42.65 $\pm$ 4.67          & \textbf{65.75 $\pm$ 2.56} & \textbf{66.44 $\pm$ 1.77} \\
                        &                         & S-06  & 22.47 $\pm$ 0.25 & 41.13 $\pm$ 3.26          & 32.35 $\pm$ 2.75          & 79.01 $\pm$ 2.72          & \textbf{91.50 $\pm$ 3.01} \\
                        &                         & S-11  & 23.70 $\pm$ 6.07 & 43.75 $\pm$ 2.85          & 54.30 $\pm$ 4.89          & \textbf{67.84 $\pm$ 2.39} & \textbf{69.98 $\pm$ 2.36} \\
                        &                         & S-12  & 32.93 $\pm$ 3.88 & 41.47 $\pm$ 2.96          & 61.57 $\pm$ 2.74          & 70.79 $\pm$ 0.90          & \textbf{81.30 $\pm$ 0.64} \\
                        \cmidrule(l){2-8} 
                        & \multirow{4}{*}{Cuboid} & S-03  & 47.00 $\pm$ 8.25 & 33.06 $\pm$ 1.83          & 60.59 $\pm$ 3.78          & 63.16 $\pm$ 1.07          & \textbf{83.75 $\pm$ 2.87} \\
                        &                         & S-04  & 51.17 $\pm$ 2.44 & 23.39 $\pm$ 4.18          & 62.28 $\pm$ 3.23          & 55.62 $\pm$ 3.35          & \textbf{71.78 $\pm$ 4.48} \\
                        &                         & S-09  & 56.84 $\pm$ 3.23 & 45.43 $\pm$ 1.33          & 58.03 $\pm$ 2.87          & \textbf{73.96 $\pm$ 3.64} & \textbf{70.12 $\pm$ 5.69} \\
                        &                         & S-10  & 60.43 $\pm$ 4.84 & 41.47 $\pm$ 3.75          & 58.58 $\pm$ 5.35          & 84.32 $\pm$ 3.11          & \textbf{94.32 $\pm$ 3.75} \\
                        \cmidrule(l){2-8} 
                        & \multirow{4}{*}{Mixed}  & S-01  & 49.22 $\pm$ 5.41 & 45.40 $\pm$ 2.48          & 61.88 $\pm$ 6.28          & 76.78 $\pm$ 1.66          & \textbf{91.83 $\pm$ 2.15} \\
                        &                         & S-02  & 30.76 $\pm$ 5.17 & 40.89 $\pm$ 4.34          & 44.76 $\pm$ 2.34          & \textbf{73.04 $\pm$ 1.26} & 64.36 $\pm$ 4.56          \\
                        &                         & S-07  & 50.71 $\pm$ 5.54 & 62.87 $\pm$ 1.73          & 71.94 $\pm$ 3.17          & \textbf{88.53 $\pm$ 1.57} & 67.76 $\pm$ 1.29          \\
                        &                         & S-08  & 30.07 $\pm$ 4.07 & 35.19 $\pm$ 2.33          & 40.38 $\pm$ 2.48          & 43.43 $\pm$ 1.30          & \textbf{53.26 $\pm$ 1.95} \\
\midrule
\multicolumn{3}{c|}{All sequences}                           & 44.04 $\pm$ 0.52 & 44.98 $\pm$ 0.44          & 56.17 $\pm$ 0.42          & 66.67 $\pm$ 0.22          & \textbf{75.07 $\pm$ 0.33}         \\ \bottomrule
\end{tabular}%
}
\end{table}

\begin{table}[]
\centering
\caption{Comparison of mean recognition accuracy (in \%) of THOR2 with end-to-end models on the OCID dataset sequences recorded using the upper camera}
\label{synocidupper}
\resizebox{\textwidth}{!}{%
\begin{tabular}{@{}ccc|ccccc@{}}
\toprule
Place                   & Scene type              & Seq. ID & RCFusion         & ViT-CNN                    & RGB-D ViT                 & THOR                      & THOR2                     \\ \midrule
\multirow{12}{*}{Table} & \multirow{4}{*}{Curved} & S-25  & 23.94 $\pm$ 1.38 & 23.94 $\pm$ 1.38          & 56.75 $\pm$ 5.59          & 49.82 $\pm$ 1.79          & \textbf{68.19 $\pm$ 3.57} \\
                        &                         & S-26  & 25.06 $\pm$ 2.10 & 25.06 $\pm$ 2.10          & 61.16 $\pm$ 4.43          & 64.48 $\pm$ 2.49          & \textbf{76.88 $\pm$ 3.22} \\
                        &                         & S-35  & 31.19 $\pm$ 3.91 & 31.19 $\pm$ 3.91          & 42.30 $\pm$ 3.64          & \textbf{58.91 $\pm$ 2.32} & \textbf{59.46 $\pm$ 2.96} \\
                        &                         & S-36  & 27.27 $\pm$ 2.25 & 27.27 $\pm$ 2.25          & \textbf{63.31 $\pm$ 1.93} & 55.67 $\pm$ 1.15          & 53.23 $\pm$ 1.56          \\
                        \cmidrule(l){2-8} 
                        & \multirow{4}{*}{Cuboid} & S-23  & 61.04 $\pm$ 5.64 & 61.04 $\pm$ 5.64          & 62.48 $\pm$ 4.07          & 66.24 $\pm$ 1.03          & \textbf{75.51 $\pm$ 3.48} \\
                        &                         & S-24  & 60.12 $\pm$ 4.72 & 60.12 $\pm$ 4.72          & 59.88 $\pm$ 0.64          & 70.38 $\pm$ 3.39          & \textbf{78.43 $\pm$ 4.18} \\
                        &                         & S-33  & 49.96 $\pm$ 3.14 & 49.96 $\pm$ 3.14          & 41.93 $\pm$ 5.75          & 61.94 $\pm$ 1.42          & \textbf{79.54 $\pm$ 4.47} \\
                        &                         & S-34  & 37.50 $\pm$ 2.85 & 37.50 $\pm$ 2.85          & 42.52 $\pm$ 4.63          & 70.22 $\pm$ 1.64          & \textbf{78.73 $\pm$ 1.67} \\
                        \cmidrule(l){2-8} 
                        & \multirow{4}{*}{Mixed}  & S-21  & 45.29 $\pm$ 4.60 & 45.29 $\pm$ 4.60          & 41.85 $\pm$ 3.43          & \textbf{73.75 $\pm$ 2.01} & \textbf{78.32 $\pm$ 5.87} \\
                        &                         & S-22  & 40.93 $\pm$ 4.88 & 40.93 $\pm$ 4.88          & \textbf{76.22 $\pm$ 6.53} & 62.54 $\pm$ 1.46          & \textbf{74.69 $\pm$ 2.06} \\
                        &                         & S-31  & 39.66 $\pm$ 3.81 & 39.66 $\pm$ 3.81          & \textbf{76.24 $\pm$ 1.58} & \textbf{75.87 $\pm$ 0.95} & \textbf{79.96 $\pm$ 4.01} \\
                        &                         & S-32  & 40.32 $\pm$ 2.46 & 40.32 $\pm$ 2.46 & \textbf{68.74 $\pm$ 3.92} & 72.47 $\pm$ 1.32          & \textbf{63.41 $\pm$ 4.67} \\
\midrule
\multirow{12}{*}{Floor} & \multirow{4}{*}{Curved} & S-05  & 20.95 $\pm$ 5.20 & 20.95 $\pm$ 5.20          & 34.31 $\pm$ 4.94          & \textbf{64.17 $\pm$ 2.94} & \textbf{68.25 $\pm$ 2.13} \\
                        &                         & S-06  & 25.78 $\pm$ 3.19 & 25.78 $\pm$ 3.19          & 43.56 $\pm$ 4.90          & 69.78 $\pm$ 3.63          & \textbf{76.71 $\pm$ 2.66} \\
                        &                         & S-11  & 23.70 $\pm$ 3.95 & 23.70 $\pm$ 3.95          & 51.45 $\pm$ 2.79          & \textbf{58.77 $\pm$ 2.74} & \textbf{63.38 $\pm$ 3.51} \\
                        &                         & S-12  & 33.94 $\pm$ 3.49 & 33.94 $\pm$ 3.49          & 58.80 $\pm$ 3.49          & 58.62 $\pm$ 1.74          & \textbf{72.07 $\pm$ 1.98} \\
                        \cmidrule(l){2-8} 
                        & \multirow{4}{*}{Cuboid} & S-03  & 37.69 $\pm$ 6.09 & 37.69 $\pm$ 6.09          & \textbf{66.07 $\pm$ 5.43} & 59.78 $\pm$ 1.07          & \textbf{71.61 $\pm$ 2.93} \\
                        &                         & S-04  & 33.26 $\pm$ 5.11 & 33.26 $\pm$ 5.11          & 51.38 $\pm$ 3.09          & 49.54 $\pm$ 1.34          & \textbf{77.22 $\pm$ 3.30} \\
                        &                         & S-09  & 57.48 $\pm$ 3.46 & \textbf{57.48 $\pm$ 3.46} & 36.21 $\pm$ 4.30          & 49.59 $\pm$ 1.35 & \textbf{60.07 $\pm$ 4.01} \\
                        &                         & S-10  & 36.20 $\pm$ 2.08 & 36.20 $\pm$ 2.08          & 45.69 $\pm$ 8.00          & \textbf{77.04 $\pm$ 2.02} & \textbf{71.17 $\pm$ 4.14} \\
                        \cmidrule(l){2-8} 
                        & \multirow{4}{*}{Mixed}  & S-01  & 38.44 $\pm$ 4.51 & 38.44 $\pm$ 4.51          & 65.02 $\pm$ 2.95          & 64.06 $\pm$ 1.16          & \textbf{85.43 $\pm$ 5.11} \\
                        &                         & S-02  & 37.46 $\pm$ 2.61 & 37.46 $\pm$ 2.61          & 51.47 $\pm$ 0.96          & \textbf{76.61 $\pm$ 1.65} & 71.56 $\pm$ 3.06          \\
                        &                         & S-07  & 44.35 $\pm$ 4.54 & 44.35 $\pm$ 4.54          & 64.15 $\pm$ 3.38          & \textbf{81.46 $\pm$ 1.78} & 56.98 $\pm$ 2.54          \\
                        &                         & S-08  & 38.44 $\pm$ 2.57 & 38.44 $\pm$ 2.57          & 30.59 $\pm$ 3.59          & \textbf{78.44 $\pm$ 2.38} & 52.89 $\pm$ 4.70\\
\midrule
\multicolumn{3}{c|}{All sequences}                           & 40.92 $\pm$ 0.28 & 46.03 $\pm$ 0.13          & 55.02 $\pm$ 0.23          & 66.50 $\pm$ 0.13          & \textbf{71.44 $\pm$ 0.15} \\ \bottomrule
\end{tabular}%
}
\end{table}

\begin{figure*}
    \centering
    \includegraphics[width=\textwidth]{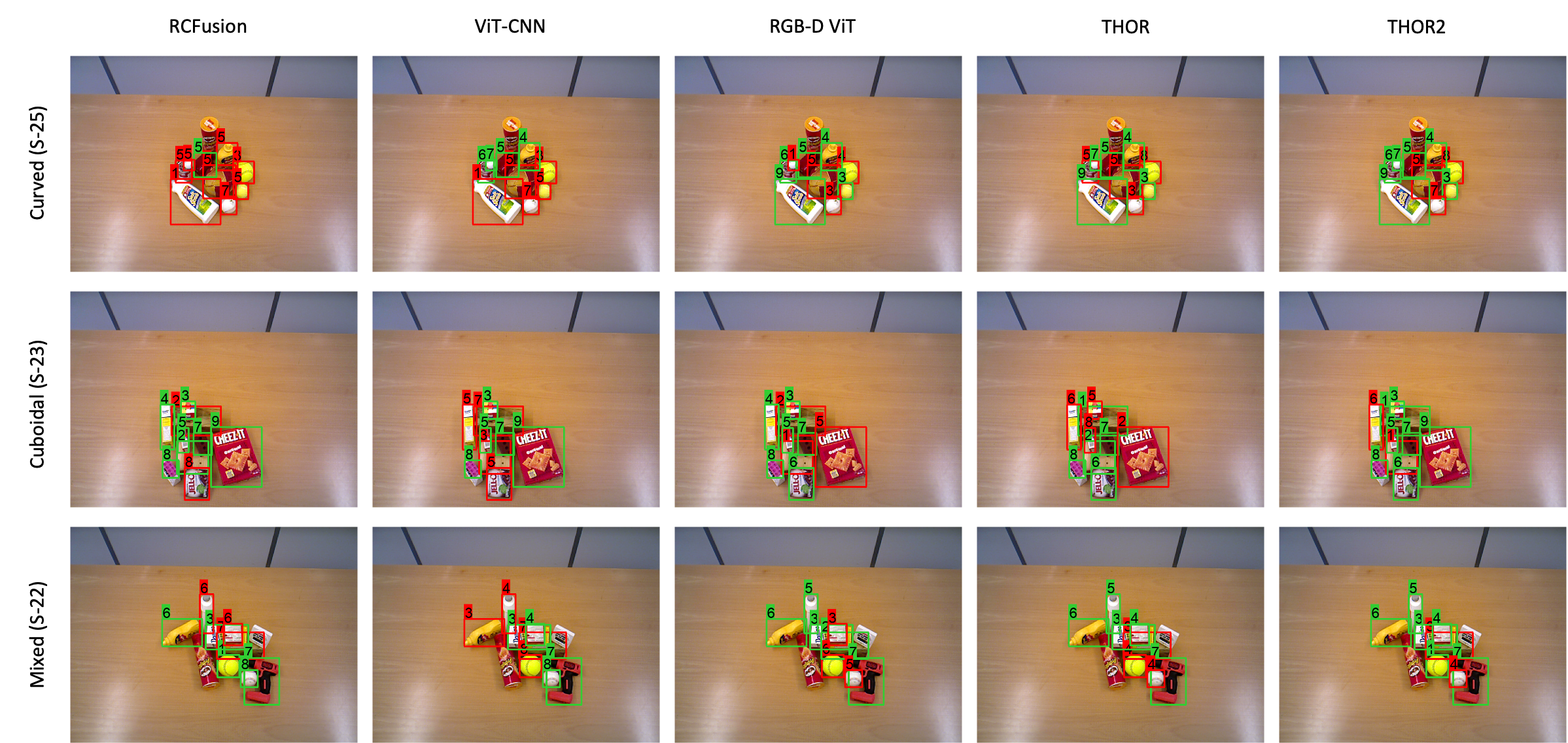}
    \caption{Sample results from the OCID dataset sequences recorded using the upper camera (green and red boxes represent correct and incorrect recognition, respectively). The three rows show results on sequences with curved, cuboidal, and mixed objects placed on a table.}
    \label{ocidresultsamplesupper}
\end{figure*}

We observe similar trends in Table \ref{synocidupper} when the test sequences are recorded using the upper camera. Although THOR2 undergoes a slight drop in performance as compared to the lower camera case, it continues to outperform THOR and RGB-D ViT by $4.94\%$ and $16.42\%$, respectively. Figure \ref{ocidresultsamplesupper} shows sample images recorded using the upper camera for the sequences that are considered in Figure \ref{ocidresultsamples}. The slight drop in THOR2's performance can be attributed to the fact that recording using the upper camera is more likely to result in challenging object poses. We refer the reader to Section \ref{discussion} for further discussion on this and other factors that affect THOR2's performance. 

\begin{figure}[h]
    \centering
    \includegraphics[width=\textwidth]{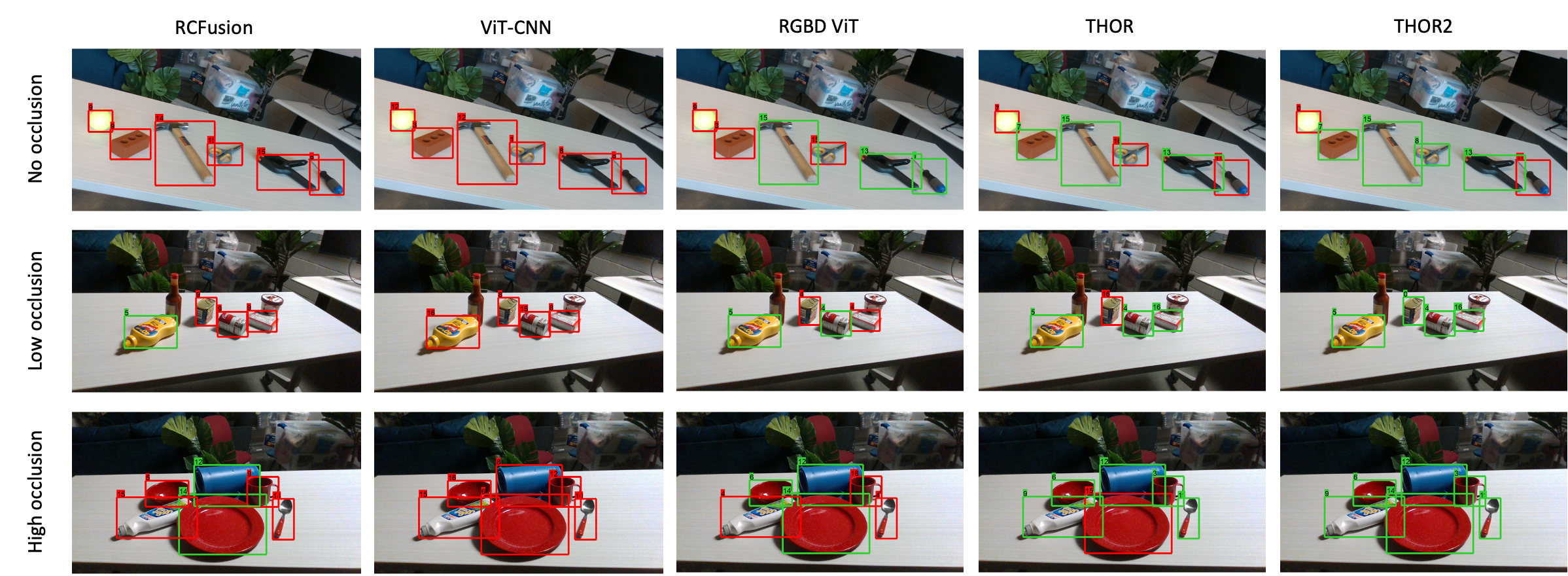}
    \caption{Sample results from the lounge environment of UW-IS Occluded dataset. The first, second, and third rows show results from scenes with three different degrees of object occlusion.}
    \label{uwis2lounge}
\end{figure}

\begin{figure}[h]
    \centering
    \includegraphics[width=\textwidth]{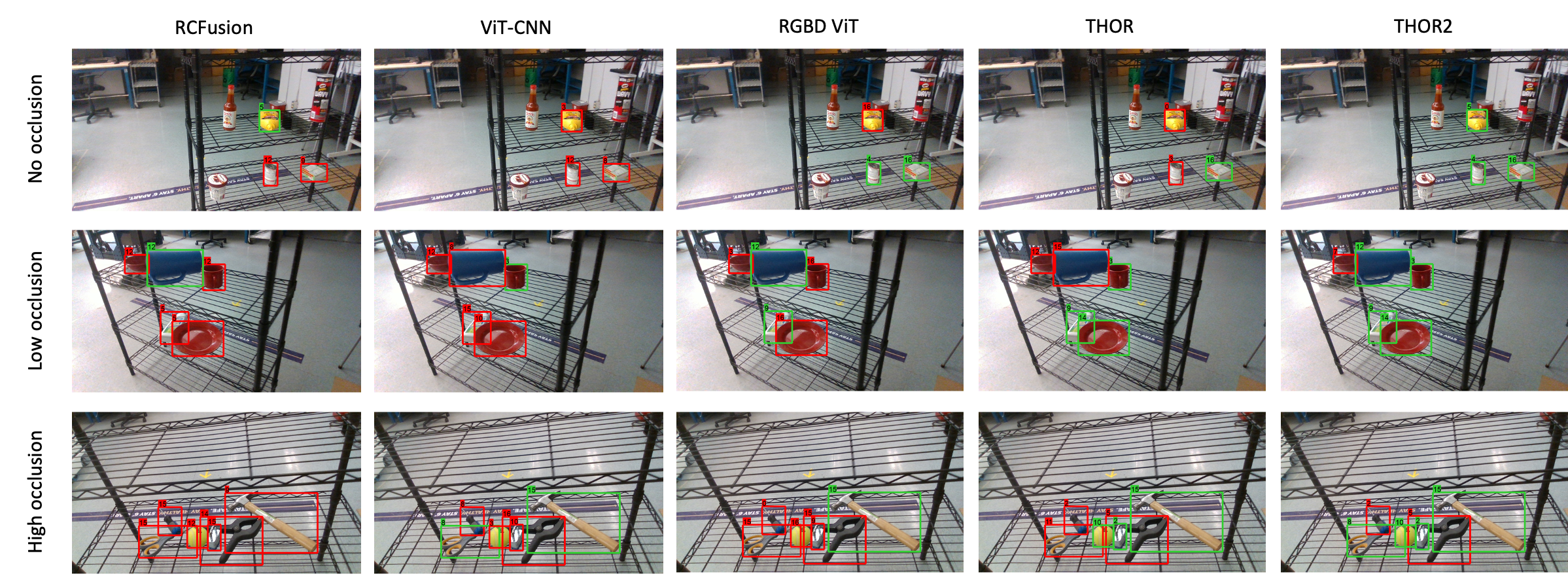}
    \caption{Sample results from the warehouse environment of UW-IS Occluded dataset. The first, second, and third rows show results from scenes with three different degrees of object occlusion.}
    \label{uwis2warehouse}
\end{figure}

\threesubsection{\textbf{Evaluation on the UW-IS Occluded dataset}}
Table \ref{uwis2objectwise}, \ref{uwis2clutterwise}, and \ref{uwis2lightingwise} show that THOR2 outperforms all the methods regardless of the environmental condition. Figure \ref{uwis2lounge} and Figure \ref{uwis2warehouse} show sample results for the different types of objects under different degrees of occlusion and lighting conditions in the lounge and warehouse environments, respectively.

In particular, Table \ref{uwis2objectwise} shows the performance in the warehouse and the lounge while considering variations in the object types. Overall, THOR2 outperforms THOR achieving approximately 10\% higher accuracy. THOR2 also achieves approximately 20\% higher accuracy than RGB-D ViT, the best-performing end-to-end model. The largest performance improvement in THOR2 (over THOR) is observed for the food category objects, some of which are similar in size and geometry but considerably distinct in color. Figure \ref{uwis2lounge} shows one such example where, unlike THOR, THOR2 successfully distinguishes between the potted meat can and the gelatin box. Relatively smaller improvements are observed for objects belonging to the tools category. We believe this observation results from the fact that several objects in this category are small, have intricate shapes, and have shiny surfaces. In such cases, the captured color information is noisy due to comparatively more inaccurate depth sensing and instance segmentation. We refer the reader to Section \ref{instancesegerrors} for further discussion on the impact of instance segmentation quality on THOR2's performance.

\begin{table}[h]
\centering
\caption{Comparison of mean recognition accuracy over object classes belonging to different types (in \%) on the UW-IS Occluded dataset}
\label{uwis2objectwise}
\resizebox{0.9\textwidth}{!}{%
\begin{tabular}{@{}cc|ccccc@{}}
\toprule
Environment & Object type & RCFusion         & ViT-CNN           & RGB-D ViT        & THOR             & THOR2                     \\ \midrule
\multirow{3}{*}{Warehouse} & Kitchen & 24.57 $\pm$ 2.79 & 18.65 $\pm$ 0.57 & 45.02 $\pm$ 1.47 & 52.33 $\pm$ 0.59 & \textbf{62.92 $\pm$ 1.34} \\
            & Tools                           & 20.15 $\pm$ 3.17 & 14.49 $\pm$ 0.80 & 35.42 $\pm$ 2.69 & 46.89 $\pm$ 0.48 & \textbf{53.55 $\pm$ 0.97} \\
            & Food                            & 16.17 $\pm$ 4.12 & 1.53 $\pm$ 0.41  & 58.31 $\pm$ 2.00 & 43.30 $\pm$ 1.48 & \textbf{65.49 $\pm$ 2.98} \\
\midrule
\multirow{3}{*}{Lounge}    & Kitchen & 24.43 $\pm$ 2.88 & 15.08 $\pm$ 0.78 & 41.71 $\pm$ 1.34 & 69.98 $\pm$ 0.43 & \textbf{78.92 $\pm$ 1.32} \\
            & Tools                           & 19.27 $\pm$ 2.26 & 11.25 $\pm$ 0.51 & 36.98 $\pm$ 1.47 & 45.96 $\pm$ 0.62 & \textbf{50.34 $\pm$ 0.96} \\
            & Food                            & 17.70 $\pm$ 3.85 & 7.75 $\pm$ 1.35  & 52.32 $\pm$ 3.74 & 45.93 $\pm$ 1.13 & \textbf{63.50 $\pm$ 3.47} \\
\midrule
\multicolumn{2}{c|}{All}              & 20.84 $\pm$ 1.17 & 12.45 $\pm$ 0.20 & 42.96 $\pm$ 0.87 & 52.22 $\pm$ 0.33 & \textbf{62.58 $\pm$ 0.36} \\ \bottomrule
\end{tabular}%
}
\end{table}

For both the environments, varying degrees of occlusion are considered in Table \ref{uwis2clutterwise}. THOR2 outperforms all the end-to-end models while gaining considerable performance improvements over THOR in the case of both unoccluded and occluded objects. Since the training data for all the methods comprise images of unoccluded objects, it is particularly interesting to note that THOR2 outperforms all the methods even in the case of unoccluded objects. We investigate this trend further, in light of the sim2real gap between training and test data, by conducting further experiments where real-world RGB-D images are used for training the best performing end-to-end model RGB-D ViT. We refer the reader to Section \ref{realworldtrainingexps} for details on those experiments. 

\begin{table}[h]
\centering
\caption{Comparison of mean recognition accuracy over all the object classes (in \%) on the UW-IS Occluded dataset under varying degrees of occlusion}
\label{uwis2clutterwise}
\resizebox{0.9\textwidth}{!}{%
\begin{tabular}{@{}cc|ccccc@{}}
\toprule
Environment & Occlusion & RCFusion        & ViT-CNN           & RGB-D ViT        & THOR             & THOR2            \\ \midrule
\multirow{3}{*}{Warehouse} & None & 20.26 $\pm$ 1.38 & 13.79 $\pm$ 0.40 & 43.65 $\pm$ 0.70 & 51.62 $\pm$ 0.53 & \textbf{61.40 $\pm$ 0.37} \\
            & Low       & 19.06 $\pm$ 1.37 & 12.77 $\pm$ 0.51 & 45.11 $\pm$ 1.37 & 48.07 $\pm$ 0.28 & \textbf{58.00 $\pm$ 0.49} \\
            & High      & 23.10 $\pm$ 1.61 & 12.25 $\pm$ 0.18 & 42.52 $\pm$ 1.07 & 44.26 $\pm$ 0.25 & \textbf{59.38 $\pm$ 0.35} \\
           \midrule
\multirow{3}{*}{Lounge}    & None & 21.43 $\pm$ 1.28 & 13.06 $\pm$ 0.23 & 39.14 $\pm$ 2.07 & 56.72 $\pm$ 0.60 & \textbf{64.29 $\pm$ 0.34} \\
            & Low       & 21.42 $\pm$ 1.36 & 9.93 $\pm$ 0.24  & 43.06 $\pm$ 0.60 & 54.45 $\pm$ 0.24 & \textbf{65.87 $\pm$ 0.64} \\
            & High      & 19.88 $\pm$ 1.43 & 13.39 $\pm$ 0.33 & 43.50 $\pm$ 1.07 & 51.88 $\pm$ 0.46 & \textbf{59.95 $\pm$ 0.52} \\ 
            \bottomrule
\end{tabular}%
}
\end{table}

Further, Table \ref{uwis2lightingwise} shows that THOR2 outperforms all the methods that use color information (i.e., RCFusion, ViT-CNN, and RGB-D ViT) regardless of the lighting conditions. It is particularly interesting to note that for each environment, the performance of THOR2 is comparable for both the lighting conditions. On the other hand, the lighting conditions impact RGB-D ViT's performance to a greater extent. This observation highlights the benefit of using the color network, which represents similar colors using a single node, to obtain the color embeddings for the TOPS2 descriptor.

\begin{table}[h]
\centering
\caption{Comparison of mean recognition accuracy over all the object classes (in \%) on the UW-IS Occluded dataset under varying lighting conditions}
\label{uwis2lightingwise}
\resizebox{0.9\textwidth}{!}{%
\begin{tabular}{@{}cc|ccccc@{}}
\toprule
Environment & Lighting & RCFusion        & ViT-CNN           & RGB-D ViT        & THOR             & THOR2            \\ \midrule
\multirow{2}{*}{Warehouse} & Bright & 20.80 $\pm$ 1.04 & 12.06 $\pm$ 0.30 & 42.15 $\pm$ 0.91 & 47.52 $\pm$ 0.36 & \textbf{61.14 $\pm$ 0.58} \\
            & Dim      & 21.12 $\pm$ 1.65 & 13.57 $\pm$ 0.35 & 47.48 $\pm$ 0.84 & 48.11 $\pm$ 0.39 & \textbf{58.28 $\pm$ 0.33} \\
            \midrule
\multirow{2}{*}{Lounge}    & Bright & 19.56 $\pm$ 0.98 & 12.46 $\pm$ 0.25 & 40.07 $\pm$ 0.79 & 50.76 $\pm$ 0.48 & \textbf{62.68 $\pm$ 0.47} \\
            & Dim      & 21.81 $\pm$ 1.45 & 11.29 $\pm$ 0.28 & 44.80 $\pm$ 1.70 & 57.64 $\pm$ 0.39 & \textbf{63.95 $\pm$ 0.51} \\ 
\bottomrule
\end{tabular}%
}
\end{table}

\subsubsection{Ablation Studies}
\label{uwis2ablationresults}

\begin{table}[h]
\centering
\caption{Comparison of mean recognition accuracy over all the object classes (in \%) achieved by using different combinations of topological representations for object recognition on the UW-IS Occluded dataset under varying degrees of occlusion.}
\label{uwis2ablation}
\resizebox{\textwidth}{!}{%
\begin{tabular}{@{}cc|ccc|ccc|c@{}}
\toprule
\multirow{2}{*}{Environment} &
  \multirow{2}{*}{Occlusion} &
  \multirow{2}{*}{CE} &
  \multirow{2}{*}{TOPS} &
  \multirow{2}{*}{TOPS2} &
  \multirow{2}{*}{CE + TOPS} &
  \multirow{2}{*}{CE + TOPS2} &
  THOR2 &
  CE + TOPS \\ 
                           &      &                  &                  &                  &                  &                  & (TOPS +   TOPS2) & + TOPS2          \\ 
                           \midrule
\multirow{3}{*}{Warehouse} & None & 48.39 $\pm$ 0.29 & 54.39 $\pm$ 0.21 & 55.08 $\pm$ 0.85 & 54.46 $\pm$ 0.45 & 51.97 $\pm$ 0.52 & \textbf{61.40 $\pm$ 0.37} & 56.09 $\pm$ 0.44 \\
                           & Low  & 46.49 $\pm$ 0.44 & 51.19 $\pm$ 0.30 & 52.07 $\pm$ 0.83 & 52.01 $\pm$ 0.48 & 49.60 $\pm$ 0.53 & \textbf{58.00 $\pm$ 0.49} & 53.23 $\pm$ 0.51 \\
                           & High & 48.93 $\pm$ 0.51 & 47.81 $\pm$ 0.24 & 55.31 $\pm$ 0.66 & 53.71 $\pm$ 0.40 & 53.35 $\pm$ 0.57 & \textbf{59.38 $\pm$ 0.35} & 56.31 $\pm$ 0.37 \\
                           \midrule
\multirow{3}{*}{Lounge}    & None & 52.43 $\pm$ 0.96 & 60.81 $\pm$ 0.09 & 55.44 $\pm$ 1.06 & 61.21 $\pm$ 0.40 & 55.43 $\pm$ 0.27 & \textbf{64.29 $\pm$ 0.34} & 61.52 $\pm$ 0.47 \\
                           & Low  & 55.39 $\pm$ 0.44 & 56.87 $\pm$ 0.29 & 59.85 $\pm$ 0.78 & 61.19 $\pm$ 0.36 & 59.32 $\pm$ 0.26 & \textbf{65.87 $\pm$ 0.64} & 63.43 $\pm$ 0.54 \\
                           & High & 49.68 $\pm$ 0.57 & 54.77 $\pm$ 0.09 & 51.07 $\pm$ 0.93 & 57.13 $\pm$ 0.29 & 52.54 $\pm$ 0.42 & \textbf{59.95 $\pm$ 0.52} & 57.95 $\pm$ 0.36 \\
                           \midrule
\multicolumn{2}{c|}{All}           & 51.01 $\pm$ 0.45 & 55.19 $\pm$ 0.12 & 55.44 $\pm$ 0.73 & 57.68 $\pm$ 0.25 & 54.58 $\pm$ 0.20 & \textbf{62.58 $\pm$ 0.36} & 59.20 $\pm$ 0.23 \\ \bottomrule
\end{tabular}%
}

{\vspace{2mm}\raggedright \underline{Note:} CE represents the case when only the vectorized color embeddings of all the slices are used as the descriptor; \texttt{nameA} + \ldots + \texttt{nameZ} indicates the case where classifiers corresponding to \texttt{nameA}, \ldots, \texttt{nameZ} are all considered during testing, and the prediction with the highest probability is selected. \par}
\end{table}

THOR2 uses two separate classifiers trained on TOPS and TOPS2 descriptors, respectively, for prediction. In this subsection, we perform ablation experiments using the UW-IS Occluded dataset to study this choice.

First, we report the recognition accuracy if only one of the descriptors is used for prediction. The fourth and fifth columns of Table \ref{uwis2ablation} report the performance when TOPS and TOPS2 are used, respectively. Interestingly, the additional color embeddings in the TOPS2 descriptor lead to minor improvements in the warehouse environment and some instances of the lounge environment. However, the overall performance of TOPS and TOPS2 descriptors on the UW-IS Occluded dataset is comparable. At the same time, the color embeddings themselves (i.e., CE) achieve a recognition accuracy approximately 8\% higher than RGB-D ViT, as shown in column three of Table \ref{uwis2ablation}. Note that CE represents the case where color embeddings are vectorized and stacked to obtain a single descriptor. We hypothesize that the fusion of color (i.e., the color embeddings) and shape information (i.e., the persistence images) in the TOPS2 descriptor provides limited flexibility in selectively using these cues during testing; more selection flexibility could lead to performance improvements. Therefore, we perform additional experiments where such flexibility is introduced by using multiple classifiers.

In particular, we obtain the recognition performance when a classifier trained on color embeddings alone is used with another classifier trained on persistence images alone (i.e., the TOPS descriptor). The results show that incorporating higher flexibility by using separate classifiers for the color embeddings and the TOPS descriptor (sixth column of Table \ref{uwis2ablation}) leads to higher overall performance than a single classifier trained using TOPS2 (fifth column of Table \ref{uwis2ablation}), where both color embeddings and persistence images are fused. 

Further analysis shows that even though selection flexibility is beneficial, the relative `importance' of the shape and color information in the framework substantially impacts the performance. For instance, using a color embeddings-based classifier and a TOPS2-based classifier (column seven of Table \ref{uwis2ablation}) implicitly assigns higher importance to color information than shape information\footnote{In this case, color information is accounted for twice, once through the color embeddings-based classifier and once through the color embeddings in the TOPS2-based classifier, whereas shape information is accounted for only once, through the TOPS2-based classifier}. Similarly, equal importance is assigned (by design) to the shape and color information in the sixth and ninth columns in Table \ref{uwis2ablation}. The only scenario where the shape information, which is a more reliable cue for recognition, is assigned higher importance than color is in the case of THOR2. As a result, THOR2 achieves higher recognition performance than all the other scenarios.

These analyses indicate that THOR2, which selectively uses the shape and color information for recognition while assigning relatively higher importance to the shape cues by design, produces the best overall recognition performance.

\subsubsection{Experiments with Real-world Training Data}
\label{realworldtrainingexps}

The previous sections report the recognition performance on real-world data when all the methods are trained using synthetic RGB-D images. By virtue of their topological nature, the TOPS and TOPS2 descriptors used in THOR2 are relatively robust to imprecise depth imagery or illumination-related changes in the real-world data. However, the other end-to-end methods are not explicitly designed to achieve robustness in such issues. Therefore, we perform additional experiments where real-world RGB-D images from the YCB dataset \cite{calli2015benchmarking} and synthetic images are used for training the end-to-end models. \revision{Table} \ref{realdatatrainingocid} \revision{reports the performance of RGB-D ViT on the OCID dataset, when it is trained using synthetic images and different amounts of real-world images from the YCB dataset. Table} \ref{realdatatraininguwis2} \revision{reports the performance of all the end-to-end methods, namely, RCFusion, ViT-CNN, and RGB-D ViT, on the UW-IS Occluded dataset, when they are trained using synthetic images and the entire YCB dataset. We note that THOR2, trained entirely on synthetic data, continues to outperform the end-to-end methods on both the OCID and UW-IS Occluded datasets.}

\begin{table}[h]
\centering
\caption{Comparison of mean recognition accuracy (in \%) on the OCID dataset sequences recorded using cameras placed at different heights. THOR2's performance is compared with RGB-D ViT trained using varying amounts of real-world data in addition to synthetic data.}
\label{realdatatrainingocid}
\resizebox{\textwidth}{!}{%
\begin{tabular}{@{}cc|c|cccc@{}}
\toprule
\multirow{2}{*}{Environment} & \multirow{2}{*}{Camera} & THOR2            & \multicolumn{4}{c}{RGB-D ViT}                                             \\ \cmidrule(l){3-7} 
                             &                         & Syn only         & Syn only         & Syn + 20\% YCB   & Syn + 60\% YCB   & Syn + 100\% YCB  \\ 
\midrule
\multirow{2}{*}{Table}       & Lower                   & \textbf{73.18 $\pm$ 0.25} & 58.02 $\pm$ 0.81 & 58.65 $\pm$ 0.24 & 63.00 $\pm$ 0.60   & 60.64 $\pm$ 0.28 \\
                             & Upper                   & \textbf{72.61 $\pm$ 0.30}  & 59.53 $\pm$ 0.32 & 57.00 $\pm$ 0.89  & 60.97 $\pm$ 0.43 & 58.83 $\pm$ 0.48 \\
\midrule
\multirow{2}{*}{Floor} & Lower & \textbf{77.09 $\pm$ 0.52} & 55.60 $\pm$ 0.38 & 54.31 $\pm$ 0.44 & 54.53 $\pm$ 0.43 & 56.98 $\pm$ 0.87 \\
                             & Upper                   & \textbf{70.10 $\pm$ 0.19}  & 51.36 $\pm$ 0.24 & 52.22 $\pm$ 0.19 & 52.89 $\pm$ 0.70  & 51.74 $\pm$ 0.49 \\
\midrule
\multirow{2}{*}{All}         & Lower                   & \textbf{75.07 $\pm$ 0.33} & 56.17 $\pm$ 0.42 & 56.36 $\pm$ 0.31 & 59.14 $\pm$ 0.27 & 57.79 $\pm$ 0.26 \\
                             & Upper                   & \textbf{71.44 $\pm$ 0.15} & 55.02 $\pm$ 0.23 & 53.73 $\pm$ 0.54 & 56.55 $\pm$ 0.22 & 54.62 $\pm$ 0.19 \\ 
\bottomrule
\end{tabular}%
}

{\vspace{2mm}\raggedright \underline{Note:} Syn + \texttt{x}\% YCB indicates that \texttt{x}\% real images from the YCB dataset are used along with the entire synthetic dataset for training and validation. \par}
\end{table}

\begin{table}[]
\centering
\caption{Comparison of mean recognition accuracy over all the object classes (in \%) on the UW-IS Occluded dataset under varying degrees of occlusion. \revision{THOR2's performance is compared with the end-to-end methods, namely, RCFusion, ViT-CNN, and RGB-D ViT trained using real-world data from the YCB dataset and synthetic data.}}
\label{realdatatraininguwis2}
\resizebox{\textwidth}{!}{%
\begin{tabular}{@{}cc|cccc@{}}
\toprule
\multirow{2}{*}{Environment}    & \multirow{2}{*}{Occlusion}    & THOR2                     & RCFusion         & ViT-CNN          & RGB-D ViT        \\ \cmidrule(l){3-6} 
                                &                               & Syn only                  & Syn + 100\% YCB        & Syn + 100\% YCB        & Syn + 100\% YCB        \\ 
                                \midrule
\multirow{3}{*}{Warehouse}          & None         & \textbf{61.40 $\pm$ 0.37} & 36.57 $\pm$ 3.07 & 14.47 $\pm$ 0.18 & 49.39 $\pm$ 2.57 \\
                                & Low                           & \textbf{58.00 $\pm$ 0.49} & 34.44 $\pm$ 3.01 & 13.30 $\pm$ 0.51 & 47.25 $\pm$ 3.01 \\
                                & High                          & \textbf{59.38 $\pm$ 0.35} & 32.57 $\pm$ 3.15 & 13.35 $\pm$ 0.37 & 46.45 $\pm$ 2.87 \\
\midrule
\multirow{3}{*}{Lounge}             & None         & \textbf{64.29 $\pm$ 0.34} & 40.37 $\pm$ 2.89 & 13.17 $\pm$ 0.45 & 46.84 $\pm$ 2.66 \\
                                & Low                           & \textbf{65.87 $\pm$ 0.64} & 35.42 $\pm$ 1.95 & 11.24 $\pm$ 0.41 & 47.51 $\pm$ 2.26 \\
                                & High                          & \textbf{59.95 $\pm$ 0.52} & 28.00 $\pm$ 2.83 & 15.41 $\pm$ 0.40 & 47.11 $\pm$ 2.90 \\
\midrule
\multicolumn{2}{c|}{Average accuracy across all   scenarios} & \textbf{62.58 $\pm$ 0.36} & 34.42 $\pm$ 2.66 & 13.47 $\pm$ 0.36 & 47.41 $\pm$ 2.70 \\
\cmidrule(r){1-2}
\multicolumn{2}{c|}{Increase in accuracy due to YCB data} & \textbf{-}                & 13.58            & 1.02             & 4.45             \\ \bottomrule
\end{tabular}%
}

{\vspace{2mm}\raggedright \underline{Note:} Syn + \texttt{x}\% YCB indicates that \texttt{x}\% real images from the YCB dataset are used along with the entire synthetic dataset for training and validation. \par}
\end{table}

The YCB dataset is recorded using five RGB and depth sensor pairs arranged in a quarter-circular arc. In one experiment, we use 20\% of the RGB-D images available for each object to train and validate the RGB-D ViT model. These images are from the first sensor pair placed at table height. In subsequent experiments, we use 60\% of the images (i.e., RGB-D images from the first three RGB and depth sensor pairs) and 100\% of the images (i.e., RGB-D images from all five RGB and depth sensor pairs placed at e). Table \ref{realdatatrainingocid} shows that, overall, a small improvement in RGB-D ViT's performance on the lower camera sequences of the OCID dataset is observed when real-world data is used for training the model. However, in the case of the upper camera sequences, the performance is somewhat lower when 20\% of the YCB dataset is used, comparable when the entire YCB dataset is used, and somewhat higher when 60\% of the dataset is used. We believe these camera-specific trends can be attributed to the difference in composition of the training set with respect to the camera viewpoints. 

\revision{In the case of the UW-IS Occluded dataset, we observe that all the end-to-end methods achieve higher recognition accuracy when real-world images from the YCB dataset are added to the training data. Overall, RGB-D ViT achieves approximately 4.5\% higher accuracy, and the highest increase is when none of the objects in the test images are occluded. This observation is expected given that the training and test data distributions are most alike in this case; both training and test data have real-world images of unoccluded objects. Similar observations can be made for RCFusion, although the gain in accuracy is substantially higher as compared to RGB-D ViT. In the case of ViT-CNN, the gain is minimal as the method does not consider retraining/fine-tuning of the feature extraction stage.} 


\revision{Overall, these results indicate that the performance of end-to-end models improves when real-world images are included during training. However, THOR2 continues to outperform them without using any real-world training data, making it particularly suitable when gathering large volumes of representative real-world training data is infeasible. We refer the reader to} Section \ref{simtoreal} \revision{for further discussion in the context of the sim2real gap between the training and test data.}



\subsection{Robot Implementation}
\label{robotimpl}
\begin{figure*}
    \centering
    \includegraphics[width=\textwidth]{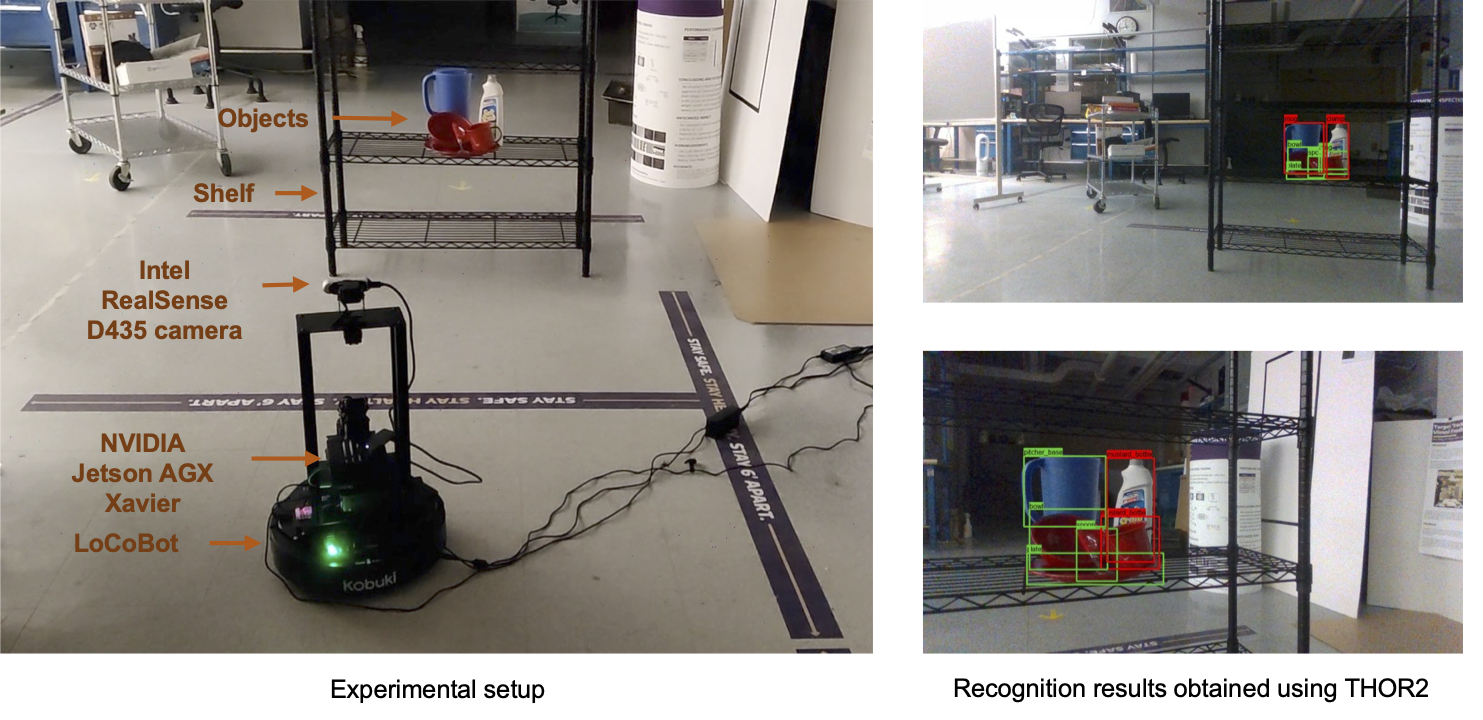}
    \caption{Screenshot of the LoCoBot operating in a mock warehouse setup (left) and the sample recognition results (right) obtained using THOR2 as the LoCoBot moves around the warehouse setup.}
    \label{robotfigure}
\end{figure*}

We also implement THOR2 on a LoCoBot platform built on a Yujin Robot Kobuki Base (YMR-K01-W1) powered by an Intel NUC NUC7i5BNH Mini PC. We mount an Intel RealSense D435 camera on top of the LoCoBot and control the robot using the PyRobot interface \cite{pyrobot2019}. The LoCoBot is equipped with NVIDIA Jetson AGX Xavier processor that has a 512-core Volta GPU with Tensor Cores and an 8-core ARM v8.2 64-bit CPU. Figure \ref{robotfigure} shows a screenshot of the platform and sample recognition results. 

THOR2 runs at an average rate of $0.7 s$ per frame in a scene with six objects on this platform. This runtime includes the time for instance segmentation that is performed using a TensorRT-optimized \cite{nvidia} depth seeding network (along with the initial mask processor module) from \cite{xie2021unseen}. Recognition prediction for every object point cloud in the scene is then obtained simultaneously using multiprocessing in Python.

\section{Discussion}
\label{discussion}
The following subsections discuss several aspects of THOR2's performance.

\subsection{THOR and THOR2 Comparison}


\label{thorvsthor2}
THOR and THOR2 differ in two key aspects. First, THOR uses 3D shape information captured in the TOPS descriptor for recognition, whereas THOR2 uses 3D shape and color information captured in the TOPS and TOPS2 descriptors. Second, THOR uses a library of trained classifiers segregated based on the number of slices and view types, whereas THOR2 only uses two classifiers, one for each descriptor. 

The results reported in Section \ref{resultsonendtoendmodels} illustrate that incorporating color information helps THOR2 achieve higher overall performance than THOR on both the OCID and the UW-IS Occluded datasets. However, unlike THOR, THOR2 undergoes a slight drop in performance when the OCID dataset sequences are recorded using the upper camera (see Table \ref{synocidlower} and \ref{synocidupper}). This observation is consistent with our expectations as THOR's viewpoint-based classifier library equips it for recognition in such scenarios where challenging object poses are more likely to occur (for e.g., S-08). 
Despite the drop in performance, THOR2 achieves higher overall accuracy than THOR. Moreover, the fewer classifiers in THOR2 make it more easily extendable to an open-world scenario where new object classes may appear. For instance, when new classes appear, a class-adaptive extension of THOR2 (with two classifiers) would be more efficiently updated than THOR, which requires updating all the classifiers in the library.








\subsection{Sim2Real Gap}
\label{simtoreal}

We train THOR2 using synthetic images of unoccluded objects and test them on real-world images of cluttered scenes. Table \ref{uwis2clutterwise} shows that THOR2 outperforms RGB-D ViT for unoccluded objects when the latter is also trained using synthetic images of unoccluded objects. Further, Table \ref{realdatatraininguwis2} shows that when synthetic and real-world RGB-D images are used to train RGB-D ViT, THOR2 continues to achieve high recognition accuracy for unoccluded objects using synthetic data alone. This observation indicates that THOR2 is better at accounting for the sim2real gap, particularly when a large volume of real-world RGB-D data is unavailable for training the RGB-D ViT. We attribute this observation to the stability of persistence images within the TOPS and TOPS2 descriptors under minor perturbations in the filtration \cite{adams2017persistence}. This stability provides some robustness against noise present in the depth images. In addition, the color network-based embeddings also provide a certain degree of robustness to the variation in object colors.

\subsection{Separability of Objects}
\revision{Separability of objects, i.e., how distinct and easily identifiable different objects are within a feature space, plays a key role in determining how challenging it can be to recognize them. Good descriptors can enhance the separability of objects, leading to improved recognition. In the case of the OCID dataset, every sequence considers a different set of objects capturing different separability scenarios. Therefore, Table} \ref{synocidlower}\revision{, Table} \ref{synocidupper}\revision{, and Table} \ref{realdatatrainingocid} \revision{consider each sequence separately and train separate models for each sequence. We also perform experiments where a single model is trained for all the sequences instead of treating each sequence individually. Table} \ref{ocid26modellower} \revision{reports recognition accuracy from these experiments for sequence recordings obtained using the lower camera. We note that the overall performance for all the methods drops compared to Table} \ref{synocidlower}\revision{, where all the sequences are considered separately. Such an outcome is expected, especially in the case of object instance recognition, because of the inherent similarities among the objects in the entire dataset. Overall, THOR2 continues to outperform all the other methods in this case as well.} 

\begin{table}[]
\centering
\caption{\revision{Comparison of mean recognition accuracy (in \%) of THOR2 with the end-to-end models on the OCID dataset sequences recorded using the lower camera (best in bold), when all the objects appearing in all the sequences are considered during training.}}
\label{ocid26modellower}
\resizebox{\textwidth}{!}{%
\begin{tabular}{@{}ccc|ccccc@{}}
\toprule
Place & Scene type & \multicolumn{1}{c|}{Seq. ID} & RCFusion & ViTCNN & RGB-D ViT & THOR & THOR2 \\ \midrule
\multirow{12}{*}{Table} & \multirow{4}{*}{Curved} & S-25 & 11.78 $\pm$ 2.76 & 15.07 $\pm$ 3.45 & 28.47 $\pm$ 5.28 & 21.78 $\pm$ 2.37 & \textbf{37.00 $\pm$ 2.06} \\
 &  & S-26 & 6.89 $\pm$ 3.69 & 9.78 $\pm$ 0.65 & 43.69 $\pm$ 8.19 & 32.42 $\pm$ 4.53 & \textbf{48.85 $\pm$ 4.12} \\
 &  & S-35 & 0.00 $\pm$ 0.00 & 9.54 $\pm$ 1.13 & 10.93 $\pm$ 3.52 & 19.10 $\pm$ 1.87 & \textbf{41.98 $\pm$ 3.26} \\
 &  & S-36 & 4.20 $\pm$ 2.33 & 20.00 $\pm$ 2.04 & \textbf{44.27 $\pm$ 7.44} & 28.12 $\pm$ 3.31 & 29.78 $\pm$ 2.69 \\
 \cmidrule(l){2-8}
 & \multirow{4}{*}{Cuboid} & S-23 & 47.04 $\pm$ 6.09 & 41.68 $\pm$ 4.19 & 53.04 $\pm$ 2.02 & 60.14 $\pm$ 3.47 & \textbf{70.66 $\pm$ 2.45} \\
 &  & S-24 & 44.01 $\pm$ 7.27 & 33.71 $\pm$ 3.84 & 46.92 $\pm$ 3.92 & 54.95 $\pm$ 1.23 & \textbf{73.73 $\pm$ 1.83} \\
 &  & S-33 & 34.17 $\pm$ 5.09 & 34.17 $\pm$ 3.77 & \textbf{57.57 $\pm$ 5.60} & \textbf{52.65 $\pm$ 2.98} & \textbf{58.99 $\pm$ 4.00} \\
 &  & S-34 & 35.43 $\pm$ 5.30 & 38.63 $\pm$ 3.97 & 58.97 $\pm$ 5.52 & 52.95 $\pm$ 1.22 & \textbf{70.95 $\pm$ 1.77} \\
 \cmidrule(l){2-8}
 & \multirow{4}{*}{Mixed} & S-21 & 27.67 $\pm$ 4.42 & 28.75 $\pm$ 2.33 & 19.47 $\pm$ 2.92 & 50.90 $\pm$ 3.84 & \textbf{60.82 $\pm$ 3.25} \\
 &  & S-22 & 12.31 $\pm$ 3.00 & 11.29 $\pm$ 1.76 & 41.88 $\pm$ 6.88 & 57.40 $\pm$ 4.81 & \textbf{64.93 $\pm$ 1.91} \\
 &  & S-31 & 30.84 $\pm$ 5.05 & 32.00 $\pm$ 1.17 & 60.89 $\pm$ 2.81 & 55.26 $\pm$ 1.76 & \textbf{70.42 $\pm$ 3.93} \\
 &  & S-32 & 25.01 $\pm$ 3.77 & 48.47 $\pm$ 2.58 & 21.31 $\pm$ 6.09 & \textbf{55.94 $\pm$ 4.29} & \textbf{60.29 $\pm$ 2.37} \\
 \midrule
\multirow{12}{*}{Floor} & \multirow{4}{*}{Curved} & S-05 & 2.22 $\pm$ 2.22 & 19.81 $\pm$ 3.99 & 16.11 $\pm$ 7.11 & 19.42 $\pm$ 4.21 & \textbf{58.77 $\pm$ 3.04} \\
 &  & S-06 & 14.89 $\pm$ 1.88 & 30.96 $\pm$ 3.55 & 21.73 $\pm$ 1.76 & 39.31 $\pm$ 4.34 & \textbf{65.86 $\pm$ 4.14} \\
 &  & S-11 & 12.44 $\pm$ 0.89 & 23.52 $\pm$ 1.43 & 29.39 $\pm$ 1.47 & 31.06 $\pm$ 5.86 & \textbf{55.72 $\pm$ 5.83} \\
 &  & S-12 & 22.58 $\pm$ 0.81 & 30.43 $\pm$ 1.47 & \textbf{49.28 $\pm$ 1.94} & 35.35 $\pm$ 2.33 & 37.66 $\pm$ 6.95 \\
 \cmidrule(l){2-8}
 & \multirow{4}{*}{Cuboid} & S-03 & 31.98 $\pm$ 2.55 & 26.15 $\pm$ 4.34 & 47.98 $\pm$ 5.54 & 59.17 $\pm$ 3.65 & \textbf{82.74 $\pm$ 2.53} \\
 &  & S-04 & 31.22 $\pm$ 3.72 & 18.55 $\pm$ 3.32 & 40.31 $\pm$ 2.58 & 59.65 $\pm$ 1.69 & \textbf{77.81 $\pm$ 1.63} \\
 &  & S-09 & 49.60 $\pm$ 2.28 & 32.00 $\pm$ 2.66 & 41.49 $\pm$ 1.71 & 56.10 $\pm$ 6.01 & \textbf{76.67 $\pm$ 2.50} \\
 &  & S-10 & 33.04 $\pm$ 4.73 & 23.91 $\pm$ 4.14 & 44.88 $\pm$ 3.93 & 71.74 $\pm$ 2.38 & \textbf{90.37 $\pm$ 2.95} \\
 \cmidrule(l){2-8}
 & \multirow{4}{*}{Mixed} & S-01 & 16.04 $\pm$ 3.56 & 39.00 $\pm$ 1.99 & 27.84 $\pm$ 4.08 & 57.23 $\pm$ 4.35 & \textbf{86.23 $\pm$ 3.74} \\
 &  & S-02 & 20.19 $\pm$ 4.36 & 25.69 $\pm$ 0.77 & 37.70 $\pm$ 0.58 & \textbf{58.26 $\pm$ 3.40} & \textbf{52.56 $\pm$ 3.90} \\
 &  & S-07 & 37.98 $\pm$ 7.42 & 45.98 $\pm$ 1.45 & \textbf{60.87 $\pm$ 3.27} & 56.21 $\pm$ 0.53 & \textbf{62.16 $\pm$ 1.93} \\
 &  & S-08 & 21.93 $\pm$ 1.92 & 25.11 $\pm$ 3.04 & \textbf{44.31 $\pm$ 3.14} & 16.19 $\pm$ 3.15 & 34.74 $\pm$ 3.89 \\
 \midrule
\multicolumn{3}{c|}{All sequences} & 29.17 $\pm$ 1.00 & 28.75 $\pm$ 0.35 & 40.03 $\pm$ 0.91 & 46.90 $\pm$ 0.68 & \textbf{59.96 $\pm$ 0.63} \\ \bottomrule
\end{tabular}%
}
\end{table}

\subsection{Challenges with THOR2}
\subsubsection{Instance Segmentation Quality}
\label{instancesegerrors}

The TOPS and TOPS2 descriptors embody the idea of object unity, making THOR2 relatively robust to over-segmentation errors or errors where the object masks are split into segments due to occlusion. However, under-segmentation and false positive segmentations are more challenging. Further, a certain degree of robustness to errors arising from misaligned object boundaries is also achieved through the test-time outlier removal from point clouds. On a related note, the temporally smoothed depth images (and point clouds) provided in the OCID dataset are relatively less noisy and do not require outlier removal. Such temporal smoothing leads to a misalignment between the RGB and depth image corresponding to a given timestep in the sequence. In sequences where the misalignment is more pronounced (e.g., sequences S-02 and S-07), the object point clouds are incorrectly colored on the boundaries resulting in a somewhat lower performance using the shape and color-based THOR2 as compared to exclusively shaped-based THOR.


\subsubsection{Specific Occlusion Scenarios}
\label{specificocclusionscenarios}
Similar to THOR, recognizing heavily occluded objects, such as the bleach cleanser in S-36, is challenging for THOR2. Further, Section \ref{trainingtesting} mentions that if an object is occluded, the aligned point cloud is reoriented to ensure that the first slice on the occluded end of the object (i.e., the end where one or more slices may be missing) is not the first slice during subsequent descriptor computation. Therefore, scenarios where an object is occluded such that there are missing slices on both the ends pose difficulties. Such occlusion scenarios are relatively infrequent, and we believe additional information, such as an RGB-D image from a different viewpoint, would be necessary to perform recognition. For instance, the bleach cleanser in S-22 appears occluded on both ends in the lower camera view (see the third row in Figure \ref{ocidresultsamples}), leading to incorrect recognition in the case of THOR and THOR2. However, the third row of Figure \ref{ocidresultsamplesupper} shows that the same bleach cleanser is correctly recognized in the upper camera view.

\section{Conclusion}

This work presents a new topological descriptor, TOPS2, and an accompanying human-inspired recognition framework, THOR2, for recognizing occluded objects in unseen indoor environments. TOPS2 is an extension of the 3D shape-based TOPS descriptor of an object's point cloud, which comprises persistence images obtained by applying persistent homology in a slicing-based manner. TOPS2 is constructed by interleaving slicing-based color embeddings that capture the point cloud's color information with the persistence images. The embeddings are computed using a network of coarse color regions obtained using the Mapper algorithm, a topological soft clustering technique. Our approach ensures similarities between the TOPS2 descriptors of the occluded and the corresponding unoccluded objects across synthetic and real-world data, embodying object unity and eliminating the need for large amounts of representative training data.

We report performance evaluation on two benchmark datasets: OCID and UW-IS Occluded. On the OCID dataset, comparisons with THOR show that THOR2 benefits from incorporating color information. Moreover, THOR2 achieves the best overall performance among all the deep learning-based end-to-end models, including the widely-used transformer, ViT, adapted for RGB-D input when increasingly cluttered scenes are viewed from different camera positions. THOR2 also achieves substantially higher recognition accuracy on the UW-IS Occluded dataset than all the methods, regardless of the environmental condition. \revision{Moreover, THOR2 continues to outperform deep learning-based end-to-end models even when real-world RGB-D images from the YCB dataset are used along with synthetic images to train the models.} 

Note that THOR2 is a passive, \textit{feedforward} approach for object recognition, i.e., a single scene image is used to recognize objects. As noted in Section \ref{specificocclusionscenarios}, heavily occluded objects can be recognized by perceiving them from a different viewpoint. Furthermore, exploring different viewpoints can also help recognize objects with ambiguous placements. Therefore, in the future, we plan to extend the THOR2 to perform active object recognition. Another promising direction of future work could be an extension of the active exploration setup where the robotic system purposefully manipulates the scene to aid its perception of the occluded objects. 



\color{black}

\medskip

%
\bibliographystyle{MSP}
\bibliography{references}

\begin{thebibliography}{10}
\providecommand{\url}[1]{\texttt{#1}}
\providecommand{\urlprefix}{URL }

\bibitem{samani2024persistent}
E.~U. Samani, A.~G. Banerjee,
\newblock \emph{IEEE Transactions on Robotics} \textbf{2024}, \emph{40} 886.

\bibitem{goldstein2016sensation}
E.~B. Goldstein, J.~Brockmole,
\newblock \emph{Sensation and perception},
\newblock Cengage Learning, \textbf{2016}.

\bibitem{macadam1942visual}
D.~L. MacAdam,
\newblock \emph{Josa} \textbf{1942}, \emph{32}, 5 247.

\bibitem{singh2007topological}
G.~Singh, F.~M{\'e}moli, G.~E. Carlsson, et~al.,
\newblock \emph{PBG@ Eurographics} \textbf{2007}, \emph{2} 091.

\bibitem{ren2015faster}
S.~Ren, K.~He, R.~Girshick, J.~Sun,
\newblock In \emph{Advances in Neural Information Processing Systems}. \textbf{2015} 91--99.

\bibitem{liu2016ssd}
W.~Liu, D.~Anguelov, D.~Erhan, C.~Szegedy, S.~Reed, C.-Y. Fu, A.~C. Berg,
\newblock In \emph{European Conference On Computer Vision}. \textbf{2016} 21--37.

\bibitem{redmon2016you}
J.~Redmon, S.~Divvala, R.~Girshick, A.~Farhadi,
\newblock In \emph{IEEE Conference on Computer Vision and Pattern Recognition}. \textbf{2016} 779--788.

\bibitem{samani2021visual}
E.~U. Samani, X.~Yang, A.~G. Banerjee,
\newblock \emph{IEEE Robotics and Automation Letters} \textbf{2021}, \emph{6}, 4 7509.

\bibitem{antonik2019human}
P.~Antonik, N.~Marsal, D.~Brunner, D.~Rontani,
\newblock \emph{Nature Machine Intelligence} \textbf{2019}, \emph{1}, 11 530.

\bibitem{lai2011large}
K.~Lai, L.~Bo, X.~Ren, D.~Fox,
\newblock In \emph{IEEE Internation Conference on Robotics and Automation}. \textbf{2011} 1817--1824.

\bibitem{kasaei2021investigating}
S.~H. Kasaei, M.~Ghorbani, J.~Schilperoort, W.~van~der Rest,
\newblock \emph{Intelligent Service Robotics} \textbf{2021}, \emph{14}, 3 329.

\bibitem{bo2013unsupervised}
L.~Bo, X.~Ren, D.~Fox,
\newblock In \emph{Experimental Robotics: The 13th International Symposium on Experimental Robotics}. \textbf{2013} 387--402.

\bibitem{gao2019rgb}
M.~Gao, J.~Jiang, G.~Zou, V.~John, Z.~Liu,
\newblock \emph{IEEE Access} \textbf{2019}, \emph{7} 43110.

\bibitem{loghmani2019recurrent}
M.~R. Loghmani, M.~Planamente, B.~Caputo, M.~Vincze,
\newblock \emph{IEEE Robotics and Automation Letters} \textbf{2019}, \emph{4}, 3 2878.

\bibitem{caglayan2022cnns}
A.~Caglayan, N.~Imamoglu, A.~B. Can, R.~Nakamura,
\newblock \emph{Computer Vision and Image Understanding} \textbf{2022}, \emph{217} 103373.

\bibitem{tziafas2023early}
G.~Tziafas, H.~Kasaei,
\newblock In \emph{IEEE/RSJ International Conference on Intelligent Robots and Systems}. \textbf{2023} 9558--9565.

\bibitem{xiong2023enhancing}
S.~Xiong, G.~Tziafas, H.~Kasaei,
\newblock In \emph{IEEE/RSJ International Conference on Intelligent Robots and Systems}. IEEE, \textbf{2023} 5751--5757.

\bibitem{corke2023robotics}
P.~Corke,
\newblock \emph{Robotics, Vision and Control: Fundamental Algorithms in Python}, volume 146,
\newblock Springer Nature, \textbf{2023}.

\bibitem{afifi2022auto}
M.~Afifi, M.~A. Brubaker, M.~S. Brown,
\newblock In \emph{Proceedings of the IEEE/CVF Winter Conference on Applications of Computer Vision}. \textbf{2022} 1210--1219.

\bibitem{paulk2014supervised}
D.~Paulk, V.~Metsis, C.~McMurrough, F.~Makedon,
\newblock In \emph{International Conference on Pervasive Technologies Related to Assistive Environments}. \textbf{2014} 1--8.

\bibitem{browatzki2011going}
B.~Browatzki, J.~Fischer, B.~Graf, H.~H. B{\"u}lthoff, C.~Wallraven,
\newblock In \emph{IEEE International Conference on Computer Vision Workshops}. \textbf{2011} 1189--1195.

\bibitem{bo2010kernel}
L.~Bo, X.~Ren, D.~Fox,
\newblock \emph{Advances in Neural Information Processing systems} \textbf{2010}, \emph{23}.

\bibitem{bo2011object}
L.~Bo, K.~Lai, X.~Ren, D.~Fox,
\newblock In \emph{IEEE Conference on Computer Vision and Pattern Recognition}. \textbf{2011} 1729--1736.

\bibitem{bo2011depth}
L.~Bo, X.~Ren, D.~Fox,
\newblock In \emph{IEEE/RSJ International Conference on Intelligent Robots and Systems}. \textbf{2011} 821--826.

\bibitem{bucak2013multiple}
S.~S. Bucak, R.~Jin, A.~K. Jain,
\newblock \emph{IEEE Transactions on Pattern Analysis and Machine Intelligence} \textbf{2013}, \emph{36}, 7 1354.

\bibitem{tuzel2006region}
O.~Tuzel, F.~Porikli, P.~Meer,
\newblock In \emph{European Conference on Computer Vision}. \textbf{2006} 589--600.

\bibitem{porikli2006covariance}
F.~Porikli, O.~Tuzel, P.~Meer,
\newblock In \emph{IEEE Computer Society Conference on Computer Vision and Pattern Recognition}, volume~1. \textbf{2006} 728--735.

\bibitem{fehr2016covariance}
D.~Fehr, W.~J. Beksi, D.~Zermas, N.~Papanikolopoulos,
\newblock \emph{Computer Vision and Image Understanding} \textbf{2016}, \emph{142} 80.

\bibitem{sun2018a}
S.~Sun, N.~An, X.~Zhao, M.~Tan,
\newblock \emph{International Journal of Advanced Robotic Systems} \textbf{2018}, \emph{15}, 1 1729881417752820.

\bibitem{blum2012learned}
M.~Blum, J.~T. Springenberg, J.~W{\"u}lfing, M.~Riedmiller,
\newblock In \emph{IEEE International Conference on Robotics and Automation}. \textbf{2012} 1298--1303.

\bibitem{cheng2015convolutional}
Y.~Cheng, R.~Cai, X.~Zhao, K.~Huang,
\newblock In \emph{International Conference on 3D Vision}. \textbf{2015} 135--143.

\bibitem{aakerberg2017depth}
A.~Aakerberg, K.~Nasrollahi, C.~B. Rasmussen, T.~B. Moeslund,
\newblock In \emph{International Joint Conference on Computational Intelligence}. \textbf{2017} 121--128.

\bibitem{eitel2015multimodal}
A.~Eitel, J.~T. Springenberg, L.~Spinello, M.~Riedmiller, W.~Burgard,
\newblock In \emph{IEEE/RSJ International Conference on Intelligent Robots and Systems}. \textbf{2015} 681--687.

\bibitem{schwarz2015rgb}
M.~Schwarz, H.~Schulz, S.~Behnke,
\newblock In \emph{IEEE International Conference on Robotics and Automation}. \textbf{2015} 1329--1335.

\bibitem{gupta2014learning}
S.~Gupta, R.~Girshick, P.~Arbel{\'a}ez, J.~Malik,
\newblock In \emph{European Conference on Computer Vision}. \textbf{2014} 345--360.

\bibitem{zaki2016convolutional}
H.~F. Zaki, F.~Shafait, A.~Mian,
\newblock In \emph{IEEE International Conference on Robotics and Automation}. \textbf{2016} 1685--1692.

\bibitem{carlucci20182}
F.~M. Carlucci, P.~Russo, B.~Caputo,
\newblock \emph{IEEE Robotics and Automation Letters} \textbf{2018}, \emph{3}, 3 2386.

\bibitem{rahman2017rgb}
M.~M. Rahman, Y.~Tan, J.~Xue, K.~Lu,
\newblock In \emph{IEEE International Conference on Multimedia and Expo}. \textbf{2017} 991--996.

\bibitem{aakerberg2017improving}
A.~Aakerberg, K.~Nasrollahi, T.~Heder,
\newblock In \emph{IEEE International Conference on Image Processing Theory, Tools and Applications}. \textbf{2017} 1--6.

\bibitem{sanchez2016comparative}
J.~Sanchez-Riera, K.-L. Hua, Y.-S. Hsiao, T.~Lim, S.~C. Hidayati, W.-H. Cheng,
\newblock \emph{Pattern Recognition Letters} \textbf{2016}, \emph{73} 1.

\bibitem{zhou2019msanet}
F.~Zhou, Y.~Hu, X.~Shen,
\newblock \emph{The Visual Computer} \textbf{2019}, \emph{35} 1583.

\bibitem{wang2015mmss}
A.~Wang, J.~Cai, J.~Lu, T.-J. Cham,
\newblock In \emph{IEEE International Conference on Computer Vision}. \textbf{2015} 1125--1133.

\bibitem{wang2015large}
A.~Wang, J.~Lu, J.~Cai, T.-J. Cham, G.~Wang,
\newblock \emph{IEEE Transactions on Multimedia} \textbf{2015}, \emph{17}, 11 1887.

\bibitem{jin2015partially}
L.~Jin, Z.~Li, X.~Shu, S.~Gao, J.~Tang,
\newblock In \emph{ACM International Conference on Multimedia}. \textbf{2015} 959--962.

\bibitem{liu2018multi}
H.~Liu, F.~Li, X.~Xu, F.~Sun,
\newblock \emph{Neurocomputing} \textbf{2018}, \emph{277} 4.

\bibitem{socher2012convolutional}
R.~Socher, B.~Huval, B.~Bath, C.~D. Manning, A.~Ng,
\newblock \emph{Advances in Neural Information Processing Systems} \textbf{2012}, \emph{25}.

\bibitem{asif2017multi}
U.~Asif, M.~Bennamoun, F.~A. Sohel,
\newblock \emph{IEEE Transactions on Pattern Analysis and Machine Intelligence} \textbf{2017}, \emph{40}, 9 2051.

\bibitem{girdhar2022omnivore}
R.~Girdhar, M.~Singh, N.~Ravi, L.~van~der Maaten, A.~Joulin, I.~Misra,
\newblock In \emph{IEEE Conference on Computer Vision and Pattern Recognition}. \textbf{2022} 16102--16112.

\bibitem{girdhar2023omnimae}
R.~Girdhar, A.~El-Nouby, M.~Singh, K.~V. Alwala, A.~Joulin, I.~Misra,
\newblock In \emph{IEEE Conference on Computer Vision and Pattern Recognition}. \textbf{2023} 10406--10417.

\bibitem{zhang2022cmx}
J.~Zhang, H.~Liu, K.~Yang, X.~Hu, R.~Liu, R.~Stiefelhagen,
\newblock \emph{arXiv preprint arXiv:2203.04838} \textbf{2022}.

\bibitem{dosovitskiy2020image}
A.~Dosovitskiy, L.~Beyer, A.~Kolesnikov, D.~Weissenborn, X.~Zhai, T.~Unterthiner, M.~Dehghani, M.~Minderer, G.~Heigold, S.~Gelly, et~al.,
\newblock \emph{arXiv preprint arXiv:2010.11929} \textbf{2020}.

\bibitem{chazal2021introduction}
F.~Chazal, B.~Michel,
\newblock \emph{Frontiers in Artificial Intelligence} \textbf{2021}, \emph{4} 667963.

\bibitem{abasi2020distance}
S.~Abasi, M.~Amani~Tehran, M.~D. Fairchild,
\newblock \emph{Color Research \& Application} \textbf{2020}, \emph{45}, 2 208.

\bibitem{luo2001development}
M.~R. Luo, G.~Cui, B.~Rigg,
\newblock \emph{Color Research \& Application: Endorsed by Inter-Society Color Council, The Colour Group (Great Britain), Canadian Society for Color, Color Science Association of Japan, Dutch Society for the Study of Color, The Swedish Colour Centre Foundation, Colour Society of Australia, Centre Fran{\c{c}}ais de la Couleur} \textbf{2001}, \emph{26}, 5 340.

\bibitem{rapp2015handbook}
B.~Rapp,
\newblock \emph{Handbook of cognitive neuropsychology: {W}hat deficits reveal about the human mind},
\newblock Psychology Press, \textbf{2015}.

\bibitem{ward2015student}
J.~Ward,
\newblock \emph{The student's guide to cognitive neuroscience},
\newblock Psychology Press, \textbf{2015}.

\bibitem{xie2021unseen}
C.~Xie, Y.~Xiang, A.~Mousavian, D.~Fox,
\newblock \emph{IEEE Transactions on Robotics} \textbf{2021}, \emph{37}, 5 1343.

\bibitem{lu2023self}
Y.~Lu, N.~Khargonkar, Z.~Xu, C.~Averill, K.~Palanisamy, K.~Hang, Y.~Guo, N.~Ruozzi, Y.~Xiang,
\newblock \emph{arXiv preprint arXiv:2302.03793} \textbf{2023}.

\bibitem{o1985finding}
J.~O'Rourke,
\newblock \emph{International journal of computer \& information sciences} \textbf{1985}, \emph{14}, 3 183.

\bibitem{suchi2019easylabel}
M.~Suchi, T.~Patten, D.~Fischinger, M.~Vincze,
\newblock In \emph{IEEE International Conference on Robotics and Automation}. \textbf{2019} 6678--6684.

\bibitem{calli2015benchmarking}
B.~Calli, A.~Walsman, A.~Singh, S.~Srinivasa, P.~Abbeel, A.~M. Dollar,
\newblock \emph{IEEE Robotics and Automation Magazine} \textbf{2015}, \emph{22}, 3 36.

\bibitem{singh2014bigbird}
A.~Singh, J.~Sha, K.~S. Narayan, T.~Achim, P.~Abbeel,
\newblock In \emph{IEEE International Conference on Robotics and Automation}. \textbf{2014} 509--516.

\bibitem{KeplerMapper_JOSS}
H.~J. van Veen, N.~Saul, D.~Eargle, S.~W. Mangham,
\newblock \emph{Journal of Open Source Software} \textbf{2019}, \emph{4}, 42 1315.

\bibitem{KeplerMapper_v1.4.1-Zenodo}
H.~J. van Veen, N.~Saul, D.~Eargle, S.~W. Mangham,
\newblock {Kepler Mapper: A flexible Python implementation of the Mapper algorithm}, \textbf{2020},
\newblock \urlprefix\url{https://doi.org/10.5281/zenodo.4077395}.

\bibitem{ester1996density}
M.~Ester, H.-P. Kriegel, J.~Sander, X.~Xu, et~al.,
\newblock In \emph{International Conference on Knowledge Discovery and Data Mining}, volume~96. \textbf{1996} 226--231.

\bibitem{panda3d_2018}
Panda3{D}: {O}pen source framework for 3{D} rendering \& {G}ames, \textbf{2018},
\newblock \urlprefix\url{https://www.panda3d.org/}.

\bibitem{yan2020depth}
C.~Yan, Z.~Li, Y.~Zhang, Y.~Liu, X.~Ji, Y.~Zhang,
\newblock \emph{ACM Transactions on Multimedia Computing, Communications, and Applications} \textbf{2020}, \emph{16}, 4 1.

\bibitem{wolf2020transformers}
T.~Wolf, L.~Debut, V.~Sanh, J.~Chaumond, C.~Delangue, A.~Moi, P.~Cistac, T.~Rault, R.~Louf, M.~Funtowicz, et~al.,
\newblock In \emph{Conference on Empirical Methods in Natural Language Processing: System Demonstrations}. \textbf{2020} 38--45.

\bibitem{pyrobot2019}
A.~Murali, T.~Chen, K.~V. Alwala, D.~Gandhi, L.~Pinto, S.~Gupta, A.~Gupta,
\newblock \emph{arXiv preprint arXiv:1906.08236} \textbf{2019}.

\bibitem{nvidia}
Nvidia,
\newblock Tensor{RT},
\newblock \urlprefix\url{https://github.com/NVIDIA/TensorRT}.

\bibitem{adams2017persistence}
H.~Adams, T.~Emerson, M.~Kirby, R.~Neville, C.~Peterson, P.~Shipman, S.~Chepushtanova, E.~Hanson, F.~Motta, L.~Ziegelmeier,
\newblock \emph{Journal of Machine Learning Research} \textbf{2017}, \emph{18}, 1 218.

\end{thebibliography}

\end{document}